\pdfoutput=1
\documentclass{article}
\usepackage{iclr2027_conference,times}

\usepackage{amsmath,amsfonts,bm}

\def\eqref#1{equation~\ref{#1}}

\def\1{\bm{1}}

\DeclareMathAlphabet{\mathsfit}{\encodingdefault}{\sfdefault}{m}{sl}
\SetMathAlphabet{\mathsfit}{bold}{\encodingdefault}{\sfdefault}{bx}{n}

\usepackage{hyperref}
\usepackage{url}
\usepackage{graphicx}
\usepackage{booktabs}
\usepackage{array}
\usepackage{multirow}
\usepackage{amsmath}
\usepackage{amssymb}
\usepackage{xspace}
\usepackage{enumitem}
\usepackage{xcolor}
\usepackage{colortbl}
\usepackage{pifont}
\usepackage{wrapfig}
\usepackage{needspace}
\usepackage[most]{tcolorbox}
\usepackage{tikz}
\usetikzlibrary{arrows.meta,positioning,calc,shapes.misc,fit,backgrounds,decorations.pathreplacing}
\usepackage{pgfplots}
\pgfplotsset{compat=1.18}
\usepackage[capitalize,nameinlink]{cleveref}
\definecolor{linkblue}{RGB}{30,60,140}\definecolor{citegreen}{RGB}{20,110,60}
\hypersetup{colorlinks=true, linkcolor=linkblue, citecolor=citegreen, urlcolor=linkblue}
\setlist[itemize]{leftmargin=*, topsep=2pt, itemsep=1pt}
\newcommand{\method}{R2S\xspace}
\newcommand{\methodlong}{Rollout-to-Scorer Distillation\xspace}
\newcommand{\student}{\pi_\theta}
\newcommand{\teacher}{T}
\newcommand{\thinkspan}{y^{\mathrm{think}}}

\newcommand{\cmark}{\ding{51}}
\newcommand{\xmark}{\ding{55}}
\definecolor{takeawaygreen}{rgb}{0.18, 0.55, 0.34}
\definecolor{box_green}{RGB}{0, 180, 139}
\definecolor{box_red}{RGB}{239, 99, 75}
\newtcolorbox{takeaway}[1]{
  enhanced,
  colback=takeawaygreen!8,
  colframe=takeawaygreen,
  boxrule=0.7pt,
  arc=5pt,
  left=3pt, right=3pt, top=2pt, bottom=2pt, boxsep=2pt,
  fontupper=\footnotesize\raggedright,
  title={#1},
  coltitle=white,
  fonttitle=\bfseries\footnotesize,
  colbacktitle=takeawaygreen!90!black}
\newcommand{\bestscore}[1]{\cellcolor{box_green!12}\textcolor{box_green}{#1}}
\newcommand{\secondscore}[1]{\cellcolor{box_red!12}\textcolor{box_red}{#1}}

\crefname{equation}{Eq.}{Eqs.}
\Crefname{equation}{Eq.}{Eqs.}

\title{Capable Teacher, Myopic Scorer:\\Reward-Free Teacher Repair for Agentic On-Policy Distillation}

\author{Shu Zhao, Tan Yu \\
NVIDIA \\
\texttt{shuzhao.cs@gmail.com}}

\iclrfinalcopy

\begin{document}

\maketitle
\lhead{Preprint}

\begin{abstract}
Agentic on-policy distillation (OPD) trains a student agent on its own trajectories while a teacher scores every token. A natural explanation of where it goes wrong is compounding error: the student's mistakes carry its trajectories beyond the region where the teacher is reliable, so recent methods keep training inside that region, through episode-length curricula, teacher-guided turns, prefix replay, or teacher intervention where the two disagree. We show this explanation is incomplete. From the states where the student departs most from the teacher, the teacher's own takeovers still finish the task, yet the student trained on its token scores keeps running out of turns. We attribute this to the myopia of next-token scoring. The teacher rates each token without seeing where the episode ends, so its scores let through the safe, delaying moves that outcome-based RL would punish with zero reward, a penalty agentic OPD lacks. We call this difference between what the teacher does and what it scores the \emph{rollout--scorer gap}. We propose \methodlong{} (\method{}), which repairs the scorer before distillation by mining high-disagreement states from the untrained student, running teacher takeovers from them, keeping the states where enough takeovers pass a reward-free check such as consensus or an LLM judge, and fine-tuning the teacher to give its takeover-informed scores on every assistant token from the divergence turn onward, without the takeover in context. The repaired teacher is frozen as the sole scorer for standard OPD, requiring no task reward during repair or training. Across three environments, two teachers and four students, \method{} outperforms three state-of-the-art agentic-OPD methods in all twelve teacher--student--environment cells, by $12.8$ success points on average over the best of them in each cell. Project page: \url{https://tree-shu-zhao.github.io/r2s/}.
\end{abstract}

\section{Introduction}
\label{sec:intro}

Agentic on-policy distillation (OPD) trains an agent on its own trajectories. The student acts in a live environment, and a stronger teacher scores every token it produces~\citep{reopd2026}. The teacher never acts in this loop; it only scores, and its next-token distribution on the student's histories is the whole training signal. This form of dense supervision also underlies the multi-teacher post-training pipelines of MiMo-V2-Flash and Nemotron-Cascade 2~\citep{mimo2026,nemotroncascade22026}.

A natural explanation of where agentic OPD goes wrong is compounding error. The student's mistakes carry its trajectories beyond the region where the teacher is reliable, and its scores there are noise. Recent agentic-OPD methods act on this reading and keep training inside the trusted region, by shortening the student's episodes~\citep{tcod2026}, handing turns to the teacher~\citep{guidedopd2026}, replaying teacher prefixes~\citep{reopd2026}, or intervening where the two disagree~\citep{futurebridge2026}, and report gains in their own settings.

\begin{figure}[t]
\centering
\includegraphics[width=\linewidth]{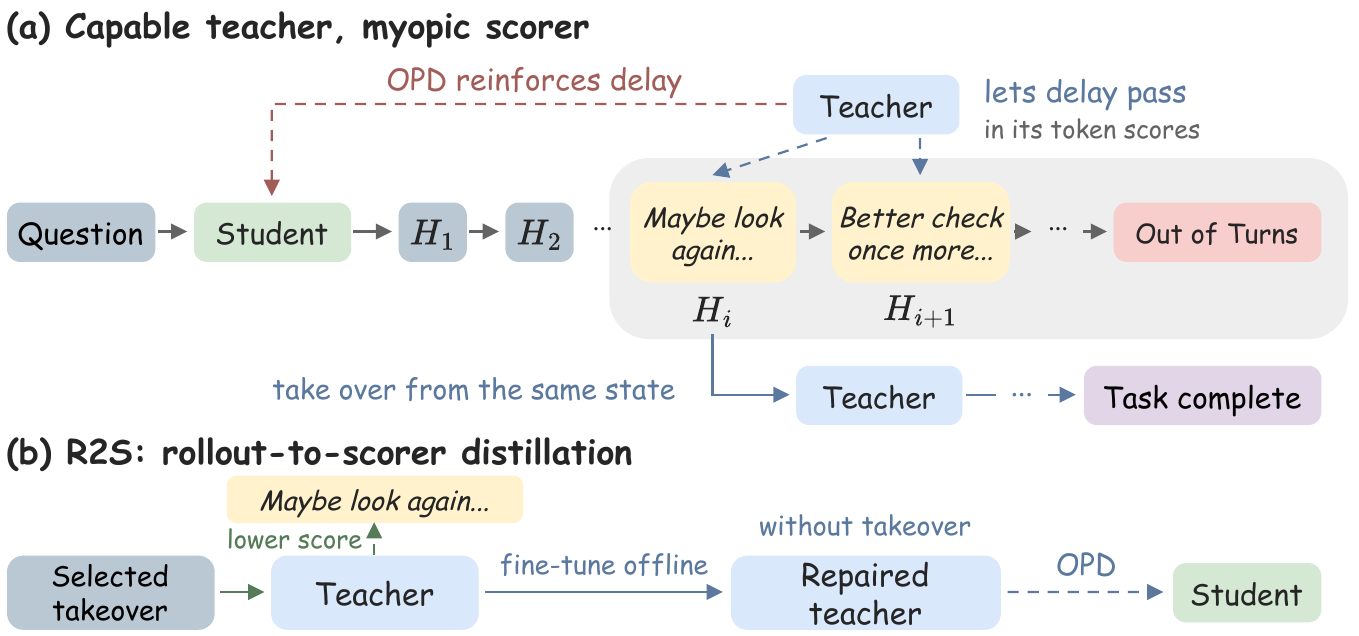}
\caption{\textbf{OPD can reinforce the delays its teacher lets pass.} (a) The teacher's scores let delaying thoughts pass and the episode runs out of turns; its own takeover from that state finishes. (b) With the takeover in context, the teacher scores the delay lower; \method{} trains a copy to do so without it.}
\label{fig:overview}
\end{figure}

We show this explanation is incomplete. We take the states where the untrained student disagrees most with the teacher, late in its episodes, the states that compounding error should place furthest outside the teacher's region, and let the teacher act from them. Its takeovers still finish the task. In every environment they beat the student's own outcome from the same states, and they rescue many of the states the student went on to lose.
The student has not drifted beyond what the teacher can handle. Yet the teacher's plain scores let the student's delay pass: only with one of its finishing takeovers in context do they mark down the student's hedging and searching, and the student trained on the plain scores keeps running out of turns. \Cref{fig:overview} illustrates this.

We attribute this to the myopia of next-token scoring. Acting, the teacher writes the whole episode itself: each action is followed by its consequences and by the teacher's own next action, and the episode runs to an outcome. Scoring, it rates each of the student's tokens one at a time, on a history it did not write, without seeing where the episode ends. Under that uncertainty another look or another search keeps options open and is reasonable on its own, so the teacher's scores let it pass. Outcome-based RL punishes such delay with zero reward when the episode runs out, while agentic OPD has no such penalty. We call this difference between what the teacher does and what it scores the \emph{rollout--scorer gap}. This motivates informing the scores with selected teacher continuations.

\methodlong{} (\method{}) repairs the scorer before distillation. First, the teacher takes over from the untrained student's highest-disagreement states. Second, we keep the states where enough takeovers pass a reward-free check, consensus among them or an LLM judge, and pick the shortest passing one as the reference. Third, we fine-tune the teacher to give its takeover-informed scores on every assistant token from the selected divergence turn onward, without the takeover in context. The repaired teacher is frozen for standard OPD. Student training retains the ordinary OPD objective and rollout recipe, and no task reward enters repair or training. We evaluate \method{} across three environments, two teachers and four students under a fixed final-checkpoint protocol. \method{} outperforms three state-of-the-art agentic-OPD methods in every cell. The gain over the best of them in each cell averages $+17.0$ points from the 9B teacher and $+8.5$ from the 30B-A3B teacher.

Our contributions are:
(1) \textbf{The rollout--scorer gap.} \Cref{sec:gap} shows that the teacher's takeovers still finish from the states where the student departs most from it, while its plain scores let the student's delay pass; only a finishing takeover in context turns them against it.
(2) \textbf{\method{}.} \Cref{sec:method} introduces a reward-free repair that moves the teacher's rollout competence into the scores OPD consumes, then freezes the teacher for standard OPD.
(3) \textbf{Evaluation.} We evaluate \method{} against three state-of-the-art agentic-OPD methods across twelve model--environment cells under a shared protocol, and measure target fidelity, out-of-turn rates and searching turns in a long-horizon analysis.

\section{Related Work}
\label{sec:related}

\textbf{On-policy distillation for agents.}
ReOPD identifies a prefix trap: histories closer to the student's distribution can make the teacher's targets less reliable, motivating a curriculum over replayed teacher prefixes~\citep{reopd2026}.
Our compared methods address related trajectory-level difficulties: TCOD-F2B schedules rollout depth, Guided-OPD allocates teacher-controlled turns, and FutureBridge validates teacher bridges from high-disagreement states~\citep{tcod2026,guidedopd2026,futurebridge2026}.
These methods retain the original scorer.
\method{} repairs its token scores before student training, a change that can be combined with student-side curricula.

\textbf{Context and self-distillation.}
SDFT and Self-Distilled Reasoner use demonstration-conditioned self-teachers to supervise student rollouts~\citep{sdft2026,opsd2026}.
SDPO instead conditions on textual feedback or successful attempts~\citep{sdpo2026}, while OPCD internalizes experiential knowledge and system prompts through on-policy context distillation~\citep{opcd2026}.
CANON selects privileged solutions by consensus without gold labels~\citep{canon2026}.
These works establish that useful distillation signals need not come from task rewards.
\method{} uses teacher takeovers to repair an external scorer, then freezes it for a separate student's OPD.

\textbf{Teacher quality and repair.}
\citet{li2026bestval} show that a higher-scoring teacher need not transfer better: compatible reasoning patterns and new capabilities matter.
\citet{rethinkingopsd2026} find that privileged context can suppress self-correction and degrade thinking models.
These results motivate checking what the teacher's local signal teaches, beyond its standalone performance.
Distillation Traps and Guards directly calibrates an external teacher with utility and calibration rewards~\citep{trapsguards2026}.
\method{} instead fits takeover-informed token targets without task reward and evaluates the repaired scorer through target fidelity and downstream completion.
\Cref{app:compare,app:related} give the detailed comparisons and extended discussion.

\section{Analysis}
\label{sec:gap}\label{sec:analysis}\label{sec:setup}\label{sec:setup-objective}\label{sec:setup-taxonomy}%
\label{sec:syndrome}\label{sec:anatomy}\label{sec:anatomy-force}\label{sec:anatomy-carrierchannel}\label{sec:anatomy-keystone}\label{sec:anatomy-lock}\label{sec:gap-general}

\subsection{Preliminaries: Agentic On-Policy Distillation}
\label{sec:prelim}

Agentic on-policy distillation (OPD) trains a student $\student$ to match a teacher $\teacher$. The student acts in the environment: at each turn $t$ it reads the history $h_t$ (the task, the observations so far and its own earlier responses) and emits a response $y_t$; the environment returns the next observation. In vanilla OPD the teacher does not act; it scores every token the student produces, given the same history.

The training signal is the teacher's next-token distribution on the student's own histories. The student minimizes a token-level divergence to it; we use the reverse KL,
\begin{equation}
\label{eq:opd}
\mathcal{L}_{\mathrm{OPD}}(\theta)=
\mathbb{E}_{h\sim\student}\;
\mathrm{KL}\big(\student(\cdot\mid h)\,\|\,\teacher(\cdot\mid h)\big),
\end{equation}
estimated at the sampled token~\citep{mimo2026,li2026bestval}. Other OPD variants use forward KL or Jensen--Shannon divergence~\citep{sdft2026,sdpo2026}; what follows concerns the per-token scoring, not the choice of divergence. The loss reports how closely the student matches the teacher's scores. Nothing in this objective looks past the current token: the teacher scores $h$ without seeing how the episode ends.
The analysis below studies one teacher, Qwen3.5-9B, and its Qwen3.5-4B student.

\subsection{Where the Student Drifts, the Teacher Still Finishes}
\label{sec:takeover}\label{sec:loss-success}

\textbf{Where the student trains is not the whole story.} Recent methods trace what goes wrong in agentic OPD to long horizons. The student's early mistakes compound~\citep{dagger2011} and carry it into states beyond the teacher's reach, where the teacher's scores can no longer guide it, so the student learns the wrong thing there. TCOD-F2B, Guided-OPD and FutureBridge therefore keep training within that reach, by shortening the student's episodes, handing turns to the teacher, or intervening where the two disagree.

The explanation assumes that the states the student drifts into lie beyond what the teacher can handle. If so, the teacher acting from them should fail as well. We test this on the states the explanation points to: the turns at which the untrained student's next-token distribution departs most from the teacher's. From each state the teacher takes over eight times. As the success columns of \Cref{tab:drift} show, in every environment its takeovers finish more often than the student did from the same states. The student has not drifted beyond what the teacher can handle.

\begin{table}[t]
\caption{\textbf{The teacher finishes from the student's states, but its plain scores let the student delay.} States: highest-disagreement turns from turn five on; \emph{Rescued}: lost states a takeover finishes. \emph{Drop}: in the teacher's probability of the student's words with a finishing takeover in context.}
\label{tab:drift}
\centering\small
\begin{tabular}{@{}lcccccc@{}}
\toprule
& \multicolumn{3}{c}{Success (\%)} & \multicolumn{3}{c}{Drop (\%)} \\
\cmidrule(lr){2-4}\cmidrule(l){5-7}
Environment & Student & Takeover & Rescued & Hedging & Searching & Other \\
\midrule\midrule
ALFWorld & $39.9$ & $44.2$ & $45.2$ & $8.2$ & $6.2$ & $3.5$ \\
WebShop & $6.1$ & $10.2$ & $14.4$ & $8.3$ & $7.7$ & $4.2$ \\
ScienceWorld & $18.6$ & $27.5$ & $51.7$ & $13.0$ & $16.6$ & $5.0$ \\
\bottomrule
\end{tabular}
\end{table}

\textbf{Its plain scores let the student's delay pass.} Once one of its finishing takeovers is in context, the teacher's scores turn against the student's delay: in the drop columns of \Cref{tab:drift}, its probability of the student's hedging and searching words falls two to three times as much as that of the rest of the student's thinking. The plain scores OPD trains on lack this direction: these states are within the teacher's reach, yet the student learns from the teacher's next-token scores there, not from its rollouts. As a result, the student trained on these plain scores to its final checkpoint still runs out of turns in a third of its held-out ALFWorld and ScienceWorld episodes and in more than half of its WebShop ones, about twice as often as when it trains on the repaired teacher.

\begin{takeaway}{Takeaway: recoverable states, cautious student}
At the states where the student departs most from the teacher, even late in its episodes, the teacher's own takeovers still finish the task. The teacher's plain scores let the student's delay pass, and the student trained on them keeps running out of turns.
\end{takeaway}

\subsection{The Scorer Never Sees Where the Episode Ends}
\label{sec:myopia}

\textbf{Outcome-based RL punishes delay with zero reward when the episode runs out of turns; plain scoring lacks this continuation signal.} Acting, the teacher generates a continuation from the student's state, observes the consequences of its actions, and reaches an outcome. Scoring, it rates the student's tokens given the observed history, without that continuation in context. Its probabilities may encode expectations about future outcomes, but ordinary OPD supplies no explicit feedback from the realized task outcome. This distinction matters when useful information gathering and costly delay look similar locally. Another look or search keeps options open, but it spends one turn of a fixed budget, which the student trained on plain scores often uses up. Plain next-token scores carry no penalty for delay: each extra look is scored on its own, and on its own it is reasonable. Conditioning the scorer on a selected teacher continuation changes its supervision on the same student text. We call the distinction between the teacher's demonstrated rollout competence and its token-level supervision the \emph{rollout--scorer gap}.

\textbf{Design implications.} We mine the untrained student's high-disagreement states and use selected teacher takeovers to inform the scores supplied there. Agreement on an observable outcome or an offline filter supplies the selection rule without reading task reward. We train the teacher to reproduce takeover-informed distributions on every assistant token from the divergence turn onward, without the takeover in context. The repaired teacher then becomes the sole scorer for ordinary OPD.

\section{Rollout-to-Scorer Distillation}
\label{sec:method}
\label{sec:repairs}
\label{sec:repairs-r2s}

\begin{figure}[t]
\centering
\includegraphics[width=\linewidth]{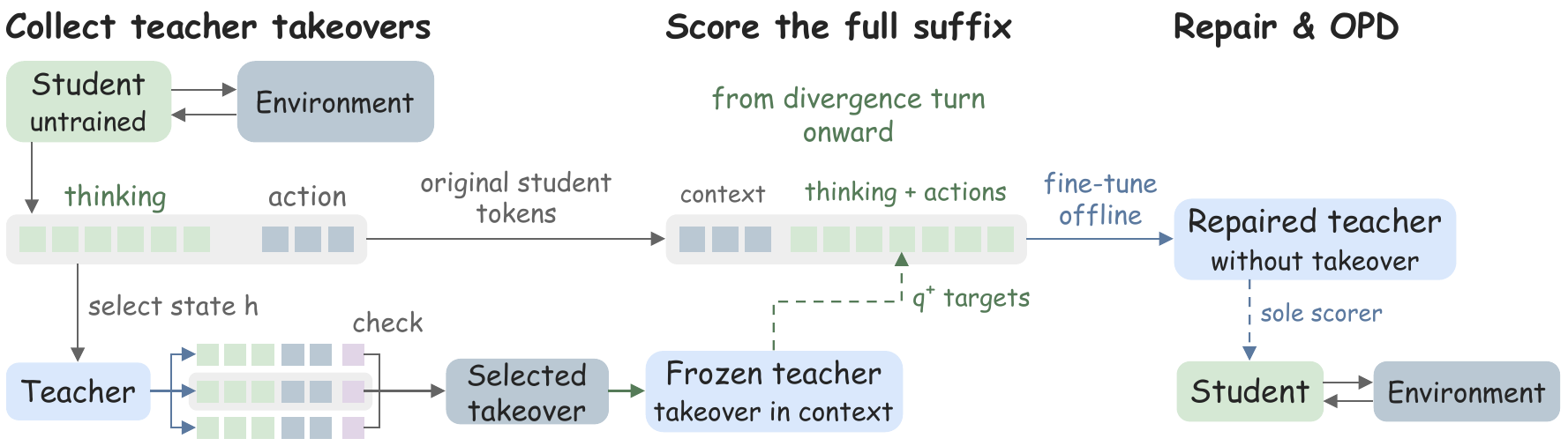}
\caption{\textbf{\method{} repairs the teacher's scores on the full assistant suffix before ordinary OPD.} With a selected takeover from the untrained student's state in context, the frozen teacher $T_0$ scores the student's thinking and actions from the divergence turn onward ($q^{+}$). A copy learns these targets without the takeover and becomes the sole OPD scorer.}
\label{fig:pipeline}
\end{figure}

\method{} targets the rollout--scorer gap of \Cref{sec:gap}. The teacher's rollouts from the student's states carry the judgment its scores lack. As illustrated in \Cref{fig:pipeline}, we therefore let the teacher take over from those states, read off the scores it gives with its own ending in view, and fine-tune it to give those scores closed-book, with no takeover in context; the refrozen teacher then scores ordinary OPD. The student never sees a takeover.

\textbf{The teacher takes over where the student departs most from it.} We first let the untrained student play the training tasks. In each of its episodes we mine the divergence turn: the turn where the student's thinking departs most from the teacher's, measured by OPD's own per-token divergence averaged over that thinking,
\begin{equation}
\label{eq:divscore}
\Delta_t=\frac{1}{|\thinkspan_t|}\sum_{i\in\thinkspan_t}
\left[\log \student(y_{t,i}\mid h_t,y_{t,<i})-\log T_0(y_{t,i}\mid h_t,y_{t,<i})\right],
\end{equation}
where $\thinkspan_t$ is the thinking of turn $t$, the part of $y_t$ before the action, and $T_0$ is the original teacher; the divergence turn has the largest $\Delta_t$. Only the thinking is compared, because it decides the action that follows. From the state at each mined turn, the teacher takes over $K$ times and plays the episode to its end. The choice among these takeovers is general: any check that tells a finishing takeover from a failing one will do, such as consensus among the takeovers, an LLM-as-a-judge, or the ground truth, which is the task's own 0--1 reward. Consensus and a judge need no reward, so \method{} can also run where no reward is available. A state is kept when at least $m$ of its takeovers pass the check, and the shortest of them is selected as $\tau_h$.

\textbf{The judgment supervises the complete suffix from the divergence state.}
For a kept state $h$, the frozen teacher $T_0$ scores the rest of the student's own episode, every token of its thinking and actions from $h$ to the end, with the takeover $\tau_h$ in context. These informed scores are the targets $q^+$. The takeover only informs them; its tokens never replace the student's. A trainable copy $T_\phi$ sees the same episode without the takeover and learns to give the same scores:
\begin{equation}
\label{eq:r2s}
\mathcal{L}_{\mathrm{R2S}}(\phi)=
\mathbb{E}_{h}\,\mathbb{E}_{h'}\Big[\mathrm{CE}\big(\underbrace{T_0(\cdot\mid\tau_h,h')}_{q^{+}},\;T_\phi(\cdot\mid h')\big)\Big],
\end{equation}
where $h'$ runs over the student's histories from $h$ to the end of its episode, one per token the student wrote, and $\mathrm{CE}$ is the cross-entropy. Refrozen, $T_\phi$ is an ordinary next-token scorer, and OPD uses it in place of $T_0$.

\section{Experiments}
\label{sec:exp}

Our experiments ask four questions. RQ1: does repairing the scorer improve OPD across teachers, students and environments? RQ2: can the takeovers be selected without reward? RQ3: which parts of \method{} are necessary? RQ4: what does the repaired scorer change in long-horizon behavior? We answer these questions below.

\begin{table}[t]
\centering
\caption{\textbf{Main results: \method{} versus vanilla OPD, TCOD-F2B, Guided-OPD and FutureBridge.} Success (\%) of one training run per cell, mean$\pm$std over three decodes of its final checkpoint; every split in \Cref{tab:main-full}. The \colorbox{box_green!12}{\textcolor{box_green}{best}} and \colorbox{box_red!12}{\textcolor{box_red}{second best}} of the five trained students in each column are highlighted.}
\label{tab:main}
\small
\setlength{\tabcolsep}{4.5pt}
\begin{tabular}{@{}lcccccc@{}}
\toprule
\multirow{3}{*}{Method} & \multicolumn{6}{c}{Qwen3.5-9B $\to$ Qwen3.5-2B\,/\,4B} \\
\cmidrule(lr){2-7}
 & \multicolumn{2}{c}{ALFWorld} & \multicolumn{2}{c}{WebShop} & \multicolumn{2}{c}{ScienceWorld} \\
\cmidrule(lr){2-3}\cmidrule(lr){4-5}\cmidrule(lr){6-7}
 & 2B & 4B & 2B & 4B & 2B & 4B \\
\midrule\midrule
Teacher alone & \multicolumn{2}{c}{$63.1{\pm}1.7$} & \multicolumn{2}{c}{$33.5{\pm}0.3$} & \multicolumn{2}{c}{$32.0{\pm}1.8$} \\
Student alone & $15.1{\pm}1.8$ & $43.4{\pm}1.0$ & $17.6{\pm}1.1$ & $33.7{\pm}0.9$ & $3.1{\pm}0.4$ & $28.0{\pm}1.5$ \\
\midrule
Vanilla OPD & $58.0{\pm}2.4$ & \secondscore{$65.8{\pm}2.6$} & $35.7{\pm}1.6$ & $34.6{\pm}2.9$ & $23.7{\pm}2.0$ & \secondscore{$34.6{\pm}3.5$} \\
TCOD-F2B & $62.7{\pm}2.0$ & $59.6{\pm}1.5$ & $33.1{\pm}0.5$ & $35.1{\pm}0.9$ & \secondscore{$28.0{\pm}2.6$} & $30.6{\pm}0.7$ \\
Guided-OPD & \secondscore{$63.6{\pm}2.2$} & $61.6{\pm}2.0$ & $35.6{\pm}0.6$ & \secondscore{$35.5{\pm}1.3$} & $25.9{\pm}3.7$ & $32.2{\pm}1.1$ \\
FutureBridge & $56.8{\pm}2.8$ & $58.4{\pm}3.7$ & \secondscore{$36.3{\pm}1.0$} & $33.9{\pm}1.7$ & $19.0{\pm}2.0$ & $30.7{\pm}1.3$ \\
\textbf{\textcolor{box_green}{\method{} (ours)}} & \bestscore{$77.9{\pm}1.3$} & \bestscore{$81.9{\pm}2.4$} & \bestscore{$57.9{\pm}1.6$} & \bestscore{$55.9{\pm}1.4$} & \bestscore{$37.9{\pm}0.6$} & \bestscore{$47.8{\pm}3.0$} \\
\midrule[\heavyrulewidth]
\multirow{3}{*}{Method} & \multicolumn{6}{c}{Qwen3-30B-A3B-Instruct-2507 $\to$ Qwen3-4B\,/\,8B} \\
\cmidrule(lr){2-7}
 & \multicolumn{2}{c}{ALFWorld} & \multicolumn{2}{c}{WebShop} & \multicolumn{2}{c}{ScienceWorld} \\
\cmidrule(lr){2-3}\cmidrule(lr){4-5}\cmidrule(lr){6-7}
 & 4B & 8B & 4B & 8B & 4B & 8B \\
\midrule\midrule
Teacher alone & \multicolumn{2}{c}{$42.3{\pm}1.1$} & \multicolumn{2}{c}{$2.9{\pm}0.3$} & \multicolumn{2}{c}{$32.9{\pm}3.2$} \\
Student alone & $30.4{\pm}2.3$ & $46.9{\pm}0.8$ & $15.5{\pm}1.2$ & $23.6{\pm}1.2$ & $9.0{\pm}1.4$ & $24.0{\pm}3.9$ \\
\midrule
Vanilla OPD & $41.2{\pm}1.7$ & $44.5{\pm}2.6$ & \secondscore{$24.7{\pm}1.1$} & $20.3{\pm}0.9$ & $38.0{\pm}1.2$ & $36.8{\pm}1.0$ \\
TCOD-F2B & \secondscore{$42.6{\pm}1.1$} & \secondscore{$48.8{\pm}1.8$} & $22.8{\pm}0.7$ & \secondscore{$23.4{\pm}0.7$} & \secondscore{$38.4{\pm}0.9$} & $34.9{\pm}0.5$ \\
Guided-OPD & $42.0{\pm}0.4$ & $47.0{\pm}1.9$ & $23.3{\pm}1.6$ & $22.3{\pm}0.8$ & $34.7{\pm}1.8$ & $33.0{\pm}1.5$ \\
FutureBridge & $41.8{\pm}2.6$ & $45.9{\pm}1.1$ & $16.7{\pm}0.9$ & $20.7{\pm}0.7$ & $36.1{\pm}1.5$ & \secondscore{$37.1{\pm}2.0$} \\
\textbf{\textcolor{box_green}{\method{} (ours)}} & \bestscore{$54.5{\pm}1.4$} & \bestscore{$54.7{\pm}1.0$} & \bestscore{$28.5{\pm}0.6$} & \bestscore{$30.4{\pm}1.2$} & \bestscore{$45.8{\pm}1.4$} & \bestscore{$50.5{\pm}1.4$} \\
\bottomrule
\end{tabular}
\end{table}

\subsection{Setup}
\label{sec:exp-setup}

\textbf{Teachers and students.}
Qwen3.5-9B~\citep{qwen352026} teaches Qwen3.5-2B and Qwen3.5-4B on three environments, and Qwen3.5-4B teaches itself in \Cref{tab:self}.
Qwen3-30B-A3B-Instruct-2507~\citep{qwen32025} teaches Qwen3-4B and Qwen3-8B on the same three environments.
We chose these pairs to vary three things: model size, from 2B to 30B parameters; model generation, Qwen3.5 and Qwen3; and architecture, a dense 9B teacher and a mixture-of-experts 30B-A3B teacher.
Every prompt carries one fixed worked demonstration per environment; native reasoning mode is off, and each turn opens with one to three sentences of plain-text deliberation before the tool call, as detailed in \Cref{app:protocol}.

\textbf{Environments.}
ALFWorld (household tasks), WebShop (shopping) and ScienceWorld (science experiments) are scored by success (\%). ALFWorld and ScienceWorld have a seen and an unseen split, and the main text reports their mean; WebShop is evaluated on its 500 canonical goals. Following TCOD, the held-out splits hold 140 seen and 134 unseen games on ALFWorld and 140 seen variations and 134 unseen task rows on ScienceWorld. An episode has 30 turns, 15 on WebShop, under one full-history tool-call protocol.

\textbf{Protocol.}
Each teacher--student--environment cell is one 150-step training run. Its final checkpoint is evaluated in three sampled decodes, with no checkpoint selection. The teacher and the untrained student acting alone follow the same prompt and evaluation protocol. In every cell, \method{} mines up to $2{,}400$ states, takes $K{=}8$ takeovers from each and fine-tunes the teacher for one epoch.

\textbf{Training and evaluation.} Each step trains on 16 episodes, 128 on WebShop, sampled at temperature 0.4 within a $32{,}768$-token context; the distillation log-ratio of the sampled estimator enters a clipped PPO surrogate as the per-token advantage, with no additional KL or entropy term, and Adam runs at a constant learning rate of $10^{-6}$. Reasoning mode is off for acting, scoring and distillation; a turn is capped at 512 tokens in training and $4{,}096$ at evaluation, and student responses are copied byte-identically into the separately served teacher context. Table entries are the mean$\pm$std over the three decodes of the final checkpoint: its in-training evaluation and two re-loads under further decoding seeds. The teacher acting alone is evaluated through the same path, prompts and seeds.

\textbf{Repair settings.} A turn is eligible for mining when its thinking has at least three tokens and ends in a tool call; we keep the eligible turn with the largest $\Delta_t$ in each episode, at most three states per goal, and the $N$ states with the largest $\Delta_t$ overall. In the privileged pass the selected takeover enters as one reference message before the student's first turn, listing the teacher's thinking and action at each of its steps; $T_\phi$ sees the same stream without that message. The targets $q^{+}$ keep the teacher's sixteen most probable tokens at each position, renormalized, and are computed once from $T_0$; $T_\phi$ is initialized from $T_0$ and fine-tuned full-parameter for one epoch at learning rate $2\times10^{-6}$, in batches of eight states, with 10\% of goals held out. \Cref{app:r2s} lists every remaining setting.

\textbf{Baselines.}
Vanilla OPD minimizes \Cref{eq:opd} with the sampled estimator against the original teacher $T_0$; \method{} runs the same OPD against the repaired teacher $T_\phi$, so vanilla OPD is its direct control.
The other three are recent agentic-OPD methods, chosen because, like \method{}, they train the student by distillation alone, without a task-reward RL objective such as GRPO~\citep{grpo2024}. TCOD-F2B~\citep{tcod2026} trains the student on shortened episodes that lengthen over training, Guided-OPD~\citep{guidedopd2026} hands part of the turns to the teacher, and FutureBridge~\citep{futurebridge2026}, the closest prior method, trains the student on validated teacher bridges from the states where the two disagree most.
Their code is public, so all three run under the same models, prompts, estimator and protocol as vanilla OPD and \method{}.
We leave out methods that combine distillation with reward-based RL, such as SDAR~\citep{sdar2026}: their students also learn from task reward, which would confound the comparison; a repaired teacher could serve as the distillation term of such a hybrid.

\subsection{Main Results}
\label{sec:exp-main}

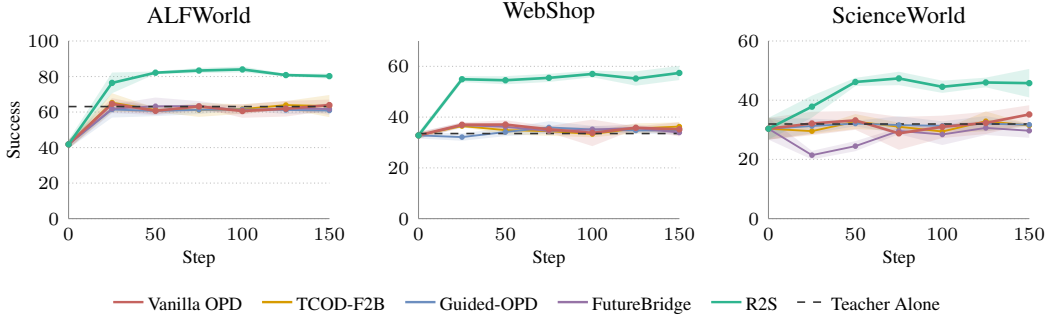
\begin{figure}[t]
\centering
\definecolor{vanRed}{HTML}{C5635F}
\definecolor{cTcod}{HTML}{D79B00}
\definecolor{cGuided}{HTML}{6C8EBF}
\definecolor{cFtb}{HTML}{9673A6}
\definecolor{r2sGreen}{HTML}{2EB58F}
\begin{tikzpicture}
\begin{axis}[
  width=3.45cm, height=2.35cm, scale only axis, xmin=0, xmax=150, ymin=0, ymax=100, xtick={0,50,100,150},
  axis lines*=left, axis line style={black!45}, tick style={draw=none}, ytick pos=left,
  ymajorgrids, grid style={densely dotted, black!25}, every tick label/.append style={font=\scriptsize},
  xlabel={Step}, xlabel style={font=\scriptsize, yshift=3pt}, ylabel={Success}, ylabel style={font=\scriptsize, yshift=-4pt},
  title={ALFWorld}, title style={font=\small, yshift=-3pt},
]
\addplot[fill=cTcod, fill opacity=0.15, draw=none, forget plot] coordinates {(0,44.26) (25,70.60) (50,62.35) (75,65.18) (100,64.30) (125,66.16) (150,69.60) (150,57.15) (125,61.70) (100,58.99) (75,57.29) (50,59.95) (25,60.00) (0,39.37)} \closedcycle;
\addplot[fill=cGuided, fill opacity=0.15, draw=none, forget plot] coordinates {(0,44.26) (25,66.00) (50,63.53) (75,63.69) (100,61.74) (125,63.55) (150,62.12) (150,59.60) (125,58.81) (100,61.19) (75,58.82) (50,57.69) (25,56.89) (0,39.37)} \closedcycle;
\addplot[fill=cFtb, fill opacity=0.15, draw=none, forget plot] coordinates {(0,44.26) (25,64.34) (50,68.17) (75,66.15) (100,63.03) (125,62.35) (150,65.00) (150,58.40) (125,61.73) (100,58.66) (75,60.58) (50,58.34) (25,59.38) (0,39.37)} \closedcycle;
\addplot[fill=vanRed, fill opacity=0.15, draw=none, forget plot] coordinates {(0,44.26) (25,67.85) (50,62.90) (75,63.80) (100,64.39) (125,66.27) (150,67.03) (150,60.82) (125,57.45) (100,56.75) (75,62.51) (50,58.53) (25,61.39) (0,39.37)} \closedcycle;
\addplot[fill=r2sGreen, fill opacity=0.15, draw=none, forget plot] coordinates {(0,44.26) (25,82.25) (50,82.87) (75,84.83) (100,85.67) (125,81.54) (150,81.65) (150,78.78) (125,80.07) (100,82.29) (75,81.89) (50,81.43) (25,70.57) (0,39.37)} \closedcycle;
\addplot[cTcod, line width=0.7pt, mark=*, mark size=0.7pt] coordinates {(0,41.82) (25,65.30) (50,61.15) (75,61.23) (100,61.65) (125,63.93) (150,63.38)};
\addplot[cGuided, line width=0.7pt, mark=*, mark size=0.7pt] coordinates {(0,41.82) (25,61.44) (50,60.61) (75,61.26) (100,61.46) (125,61.18) (150,60.86)};
\addplot[cFtb, line width=0.7pt, mark=*, mark size=0.7pt] coordinates {(0,41.82) (25,61.86) (50,63.25) (75,63.37) (100,60.85) (125,62.04) (150,61.70)};
\addplot[vanRed, line width=1.0pt, mark=*, mark size=0.7pt] coordinates {(0,41.82) (25,64.62) (50,60.72) (75,63.16) (100,60.57) (125,61.86) (150,63.92)};
\addplot[r2sGreen, line width=1.0pt, mark=*, mark size=0.7pt] coordinates {(0,41.82) (25,76.41) (50,82.15) (75,83.36) (100,83.98) (125,80.81) (150,80.21)};
\draw[dashed, black!80, line width=0.6pt] (axis cs:0,63.08) -- (axis cs:150,63.08);
\end{axis}
\end{tikzpicture}
\hfill
\begin{tikzpicture}
\begin{axis}[
  width=3.45cm, height=2.35cm, scale only axis, xmin=0, xmax=150, ymin=0, ymax=70, xtick={0,50,100,150},
  axis lines*=left, axis line style={black!45}, tick style={draw=none}, ytick pos=left,
  ymajorgrids, grid style={densely dotted, black!25}, every tick label/.append style={font=\scriptsize},
  xlabel={Step}, xlabel style={font=\scriptsize, yshift=3pt}, ylabel={}, ylabel style={font=\scriptsize, yshift=-4pt},
  title={WebShop}, title style={font=\small, yshift=-3pt},
]
\addplot[fill=cTcod, fill opacity=0.15, draw=none, forget plot] coordinates {(0,34.05) (25,36.88) (50,36.57) (75,36.31) (100,34.62) (125,37.54) (150,37.93) (150,34.60) (125,33.53) (100,32.05) (75,32.62) (50,33.17) (25,36.05) (0,31.55)} \closedcycle;
\addplot[fill=cGuided, fill opacity=0.15, draw=none, forget plot] coordinates {(0,34.05) (25,33.58) (50,35.76) (75,38.40) (100,36.55) (125,35.91) (150,36.54) (150,33.06) (125,33.56) (100,33.71) (75,33.46) (50,32.50) (25,30.69) (0,31.55)} \closedcycle;
\addplot[fill=cFtb, fill opacity=0.15, draw=none, forget plot] coordinates {(0,34.05) (25,37.86) (50,37.00) (75,35.69) (100,36.46) (125,37.17) (150,34.44) (150,33.43) (125,34.17) (100,33.80) (75,33.38) (50,35.40) (25,35.74) (0,31.55)} \closedcycle;
\addplot[fill=vanRed, fill opacity=0.15, draw=none, forget plot] coordinates {(0,34.05) (25,37.92) (50,38.77) (75,36.45) (100,39.04) (125,36.33) (150,38.02) (150,32.38) (125,35.27) (100,28.56) (75,33.95) (50,35.37) (25,36.08) (0,31.55)} \closedcycle;
\addplot[fill=r2sGreen, fill opacity=0.15, draw=none, forget plot] coordinates {(0,34.05) (25,56.11) (50,56.15) (75,57.08) (100,58.31) (125,58.02) (150,60.20) (150,54.60) (125,52.38) (100,55.69) (75,53.85) (50,52.92) (25,53.76) (0,31.55)} \closedcycle;
\addplot[cTcod, line width=0.7pt, mark=*, mark size=0.7pt] coordinates {(0,32.80) (25,36.47) (50,34.87) (75,34.47) (100,33.33) (125,35.53) (150,36.27)};
\addplot[cGuided, line width=0.7pt, mark=*, mark size=0.7pt] coordinates {(0,32.80) (25,32.13) (50,34.13) (75,35.93) (100,35.13) (125,34.73) (150,34.80)};
\addplot[cFtb, line width=0.7pt, mark=*, mark size=0.7pt] coordinates {(0,32.80) (25,36.80) (50,36.20) (75,34.53) (100,35.13) (125,35.67) (150,33.93)};
\addplot[vanRed, line width=1.0pt, mark=*, mark size=0.7pt] coordinates {(0,32.80) (25,37.00) (50,37.07) (75,35.20) (100,33.80) (125,35.80) (150,35.20)};
\addplot[r2sGreen, line width=1.0pt, mark=*, mark size=0.7pt] coordinates {(0,32.80) (25,54.93) (50,54.53) (75,55.47) (100,57.00) (125,55.20) (150,57.40)};
\draw[dashed, black!80, line width=0.6pt] (axis cs:0,33.53) -- (axis cs:150,33.53);
\end{axis}
\end{tikzpicture}
\hfill
\begin{tikzpicture}
\begin{axis}[
  width=3.45cm, height=2.35cm, scale only axis, xmin=0, xmax=150, ymin=0, ymax=60, xtick={0,50,100,150},
  axis lines*=left, axis line style={black!45}, tick style={draw=none}, ytick pos=left,
  ymajorgrids, grid style={densely dotted, black!25}, every tick label/.append style={font=\scriptsize},
  xlabel={Step}, xlabel style={font=\scriptsize, yshift=3pt}, ylabel={}, ylabel style={font=\scriptsize, yshift=-4pt},
  title={ScienceWorld}, title style={font=\small, yshift=-3pt},
]
\addplot[fill=cTcod, fill opacity=0.15, draw=none, forget plot] coordinates {(0,34.07) (25,30.58) (50,35.31) (75,32.40) (100,30.15) (125,36.23) (150,31.92) (150,31.00) (125,29.56) (100,28.70) (75,29.70) (50,30.24) (25,28.53) (0,26.68)} \closedcycle;
\addplot[fill=cGuided, fill opacity=0.15, draw=none, forget plot] coordinates {(0,34.07) (25,33.78) (50,33.06) (75,34.58) (100,32.87) (125,34.65) (150,32.01) (150,31.41) (125,29.51) (100,29.79) (75,28.66) (50,31.17) (25,28.78) (0,26.68)} \closedcycle;
\addplot[fill=cFtb, fill opacity=0.15, draw=none, forget plot] coordinates {(0,34.07) (25,22.89) (50,26.19) (75,30.73) (100,32.00) (125,33.05) (150,32.07) (150,27.31) (125,28.25) (100,24.86) (75,28.51) (50,22.78) (25,19.99) (0,26.68)} \closedcycle;
\addplot[fill=vanRed, fill opacity=0.15, draw=none, forget plot] coordinates {(0,34.07) (25,36.11) (50,36.36) (75,34.38) (100,34.78) (125,36.07) (150,38.31) (150,32.07) (125,28.58) (100,27.21) (75,23.24) (50,30.01) (25,28.18) (0,26.68)} \closedcycle;
\addplot[fill=r2sGreen, fill opacity=0.15, draw=none, forget plot] coordinates {(0,34.07) (25,41.55) (50,47.55) (75,49.63) (100,46.60) (125,47.62) (150,50.57) (150,40.94) (125,44.36) (100,42.46) (75,45.08) (50,44.77) (25,34.09) (0,26.68)} \closedcycle;
\addplot[cTcod, line width=0.7pt, mark=*, mark size=0.7pt] coordinates {(0,30.37) (25,29.55) (50,32.77) (75,31.05) (100,29.42) (125,32.90) (150,31.46)};
\addplot[cGuided, line width=0.7pt, mark=*, mark size=0.7pt] coordinates {(0,30.37) (25,31.28) (50,32.12) (75,31.62) (100,31.33) (125,32.08) (150,31.71)};
\addplot[cFtb, line width=0.7pt, mark=*, mark size=0.7pt] coordinates {(0,30.37) (25,21.44) (50,24.49) (75,29.62) (100,28.43) (125,30.65) (150,29.69)};
\addplot[vanRed, line width=1.0pt, mark=*, mark size=0.7pt] coordinates {(0,30.37) (25,32.14) (50,33.18) (75,28.81) (100,31.00) (125,32.33) (150,35.19)};
\addplot[r2sGreen, line width=1.0pt, mark=*, mark size=0.7pt] coordinates {(0,30.37) (25,37.82) (50,46.16) (75,47.36) (100,44.53) (125,45.99) (150,45.76)};
\draw[dashed, black!80, line width=0.6pt] (axis cs:0,31.96) -- (axis cs:150,31.96);
\end{axis}
\end{tikzpicture}
\\[2pt]
{\scriptsize \tikz[baseline=-0.4ex]{\draw[vanRed, line width=0.9pt] (0,0.065) -- (0.4,0.065);}\,Vanilla OPD\quad\tikz[baseline=-0.4ex]{\draw[cTcod, line width=0.9pt] (0,0.065) -- (0.4,0.065);}\,TCOD-F2B\quad\tikz[baseline=-0.4ex]{\draw[cGuided, line width=0.9pt] (0,0.065) -- (0.4,0.065);}\,Guided-OPD\quad\tikz[baseline=-0.4ex]{\draw[cFtb, line width=0.9pt] (0,0.065) -- (0.4,0.065);}\,FutureBridge\quad\tikz[baseline=-0.4ex]{\draw[r2sGreen, line width=0.9pt] (0,0.065) -- (0.4,0.065);}\,\method{}\quad\tikz[baseline=-0.4ex]{\draw[dashed, black!80, line width=0.6pt] (0,0.065) -- (0.4,0.065);}\,Teacher Alone}
\caption{\textbf{\method{} students separate from every baseline during training.} Qwen3.5-9B teaches Qwen3.5-4B; mean $\pm$ one standard deviation over three training runs; dashed: the teacher acting alone. Other teacher--student pairs are in \Cref{app:curves}.}
\label{fig:teacher-bars}\label{fig:curves}
\end{figure}

\textbf{\method{} beats every baseline in all twelve cells of \Cref{tab:main} and ends above the teacher acting alone.} The gain over vanilla OPD averages $+17.8$ points from the 9B teacher and $+9.8$ from the 30B-A3B teacher, and \method{} exceeds every baseline in mean on all $24$ splits of \Cref{tab:main-full}.
Averaged over the ten cells outside the 30B-A3B teacher's WebShop pair, the vanilla student ends within a point of the teacher acting alone. On WebShop the 30B-A3B teacher acting alone seldom completes a purchase: on $13.7$ of its $15$ turns per episode, on average, it answers without calling the action tool, so $94\%$ of its episodes run out of turns, and every student ends above it. \Cref{fig:teacher-bars} follows the Qwen3.5-9B-to-4B students of all five methods over three training runs each: the \method{} student separates from every baseline well before the final checkpoint. The three \method{} runs share the one repaired teacher of \Cref{tab:main}, so the width of its band comes from student training alone. The teacher's takeovers from the untrained student's highest-disagreement states finish more often than the student does in eleven of the twelve cells (\Cref{tab:takeover-all} in \Cref{app:analysis}), so what \Cref{sec:takeover} found for one pair holds for all four.

\begin{table}[h]
\centering
\caption{\textbf{Self-distillation: Qwen3.5-4B teaches Qwen3.5-4B.} \method{} first repairs the 4B teacher on its own takeovers; the other arms learn from the unrepaired 4B.}
\label{tab:self}
\small
\setlength{\tabcolsep}{4.5pt}
\begin{tabular}{@{}lcccccc@{}}
\toprule
Environment & Student alone & Vanilla OPD & TCOD-F2B & Guided-OPD & FutureBridge & \textbf{\textcolor{box_green}{\method{} (ours)}} \\
\midrule\midrule
ALFWorld & $43.4{\pm}1.0$ & $49.9{\pm}2.6$ & $49.7{\pm}2.6$ & $48.5{\pm}3.8$ & \secondscore{$50.6{\pm}1.9$} & \bestscore{$72.7{\pm}1.3$} \\
WebShop & $33.7{\pm}0.9$ & $33.3{\pm}2.2$ & $31.3{\pm}0.9$ & \secondscore{$34.0{\pm}1.8$} & $31.1{\pm}1.5$ & \bestscore{$58.1{\pm}2.5$} \\
ScienceWorld & $28.0{\pm}1.5$ & $28.2{\pm}2.3$ & $27.2{\pm}0.8$ & \secondscore{$29.3{\pm}1.8$} & $29.2{\pm}2.9$ & \bestscore{$42.4{\pm}3.4$} \\
\bottomrule
\end{tabular}
\end{table}

\textbf{The teacher can be the student's own base model.}
\Cref{tab:self} repeats the Qwen3.5-4B cells with the 4B as its own teacher. In all three environments, the \method{} student taught by the repaired 4B passes the vanilla student that the 9B teaches in \Cref{tab:main}. Since the teacher and the student are the same model, the student--teacher gap that ranks states is zero, so the pipeline draws one turn per episode of the untrained student at random under the same goal cap and budget; the 4B itself runs the takeovers and the selection, so no larger model enters the pipeline. In all three environments this self-taught student also ends above the 9B teacher acting alone in \Cref{tab:main}. Taught by the unrepaired 4B, the four other students gain about six points on ALFWorld and at most about a point on WebShop and ScienceWorld: a model's plain scores of its own tokens teach it little. What lifts the student is not a larger model's knowledge but scores informed by the ending, which its own base model can supply.

\Needspace{17\baselineskip}
\subsection{Ablations}
\label{sec:exp-repair}

\begin{wraptable}{r}{0.5\textwidth}
\vspace{-10pt}
\centering
\caption{\raggedright Ablation on \textbf{takeover selection}. Success (\%). Ground truth reads the task reward and is not ranked.}
\label{tab:selection}
\fontsize{8}{9.5}\selectfont
\setlength{\tabcolsep}{4pt}
\begin{tabular}{@{}lccc@{}}
\toprule
& \textbf{ALFWorld} & \textbf{WebShop} & \textbf{ScienceWorld} \\
\midrule\midrule
Vanilla OPD & $65.8{\pm}2.6$ & $34.6{\pm}2.9$ & $34.6{\pm}3.5$ \\
Consensus vote & \secondscore{$81.9{\pm}2.4$} & \bestscore{$63.1{\pm}1.6$} & \secondscore{$39.5{\pm}2.9$} \\
LLM judge & \bestscore{$82.6{\pm}0.9$} & \secondscore{$55.9{\pm}1.4$} & \bestscore{$50.3{\pm}1.5$} \\
\midrule
Ground truth & $85.9{\pm}1.2$ & $65.0{\pm}1.4$ & $49.0{\pm}1.1$ \\
\bottomrule
\end{tabular}
\vspace{-9pt}
\end{wraptable}

Each ablation below varies one part of \method{} in the Qwen3.5-9B-to-4B cells of \Cref{tab:main}, on all three environments.

\textbf{Takeover selection.}\label{sec:exp-select}
\Cref{tab:selection} repairs the teacher with three rules for picking the winning takeovers. Ground truth reads the task reward; the consensus vote and the LLM judge do not. With either reward-free rule the student ends above vanilla OPD in all three environments, so the repair depends neither on reading the reward nor on one particular rule.

\Needspace{14\baselineskip}
\begin{wraptable}{r}{0.55\textwidth}
\vspace{-10pt}
\centering
\caption{\raggedright Ablation on \textbf{components}. Success (\%); the rows are vanilla OPD, label-SFT and \method{}.}
\label{tab:abl-components}
\fontsize{8}{9.5}\selectfont
\setlength{\tabcolsep}{3pt}
\begin{tabular}{@{}cc|ccc@{}}
\toprule
\textbf{Takeovers} & \shortstack{\textbf{Informed}\\\textbf{targets}} & \textbf{ALFWorld} & \textbf{WebShop} & \textbf{ScienceWorld} \\
\midrule\midrule
\textcolor{gray}{\xmark} & \textcolor{gray}{\xmark} & $65.8{\pm}2.6$ & $34.6{\pm}2.9$ & $34.6{\pm}3.5$ \\
\textcolor{box_green}{\cmark} & \textcolor{gray}{\xmark} & \secondscore{$76.8{\pm}3.5$} & \secondscore{$46.7{\pm}1.3$} & \secondscore{$41.0{\pm}3.0$} \\
\midrule
\textcolor{box_green}{\cmark} & \textcolor{box_green}{\cmark} & \bestscore{$81.9{\pm}2.4$} & \bestscore{$55.9{\pm}1.4$} & \bestscore{$47.8{\pm}3.0$} \\
\bottomrule
\end{tabular}
\vspace{-8pt}
\end{wraptable}

\textbf{Effect of informed targets.}
\Cref{tab:abl-components} separates the takeovers from the informed targets built on them. The label-SFT control fine-tunes the teacher on the selected takeovers by standard SFT. Its student already ends well above vanilla OPD in all three environments, so the takeovers carry signal of their own; fitting the informed targets instead lifts the student further. The takeovers help most as context for the teacher's scores of the student's own tokens, not as SFT data.

\Needspace{13\baselineskip}
\begin{wraptable}{r}{0.55\textwidth}
\vspace{-10pt}
\centering
\caption{\raggedright Ablation on \textbf{supervised turns}. Success (\%); the suffix, from the divergence turn to the end, is \method{}.}
\label{tab:abl-scope}
\fontsize{8}{9.5}\selectfont
\setlength{\tabcolsep}{3pt}
\begin{tabular}{@{}lcccc@{}}
\toprule
& \textbf{ALFWorld} & \textbf{WebShop} & \textbf{ScienceWorld} & \textbf{Avg.} \\
\midrule\midrule
Divergence turn & \bestscore{$82.5{\pm}2.3$} & $48.5{\pm}1.4$ & \bestscore{$48.7{\pm}0.4$} & 59.9 \\
Prefix & $81.0{\pm}1.8$ & \secondscore{$52.7{\pm}0.8$} & $47.6{\pm}2.1$ & \secondscore{60.5} \\
Suffix & \secondscore{$81.9{\pm}2.4$} & \bestscore{$55.9{\pm}1.4$} & \secondscore{$47.8{\pm}3.0$} & \bestscore{61.9} \\
\bottomrule
\end{tabular}
\end{wraptable}

\textbf{Supervised turns.}
\method{} fits the informed targets on the suffix of the student's episode, from the divergence turn to its end. \Cref{tab:abl-scope} supervises instead the divergence turn alone or the prefix up to it. Every span lifts the student well above vanilla OPD in all three environments, so the informed targets help whichever turns they cover. Vanilla OPD, whose teacher scores each token without seeing how the episode ends, trails them all by a wide margin.

\subsection{Long-Horizon Behavior}
\label{sec:exp-longhorizon}

We compare the Qwen3.5-9B-to-4B \method{} teachers and students of \Cref{tab:main} with vanilla OPD.

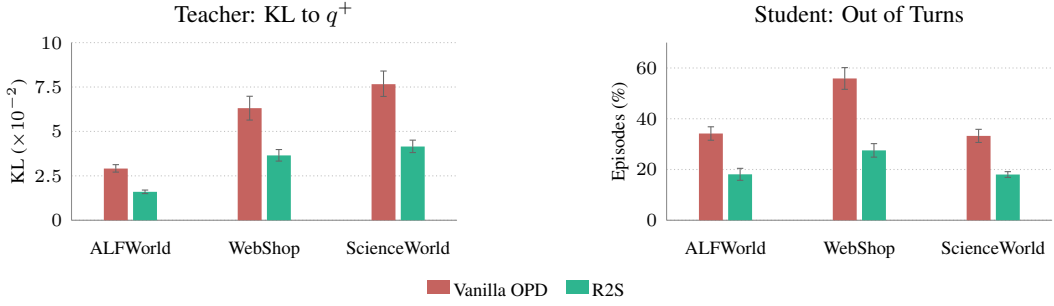
\begin{figure}[t]
\centering
\definecolor{vanRed}{HTML}{C5635F}
\definecolor{r2sGreen}{HTML}{2EB58F}
\begin{tikzpicture}
\begin{axis}[
  width=5.1cm, height=2.35cm, scale only axis, ybar, bar width=9pt, bar shift auto, enlarge x limits=0.22,
  symbolic x coords={ALFWorld,WebShop,ScienceWorld}, xtick=data, ymin=0, ymax=10, ytick={0,2.5,5,7.5,10},
  axis lines*=left, axis line style={black!45}, tick style={draw=none}, ytick pos=left,
  ymajorgrids, grid style={densely dotted, black!25}, every tick label/.append style={font=\scriptsize},
  ylabel={KL ($\times 10^{-2}$)}, ylabel style={font=\scriptsize, yshift=-4pt},
  title={Teacher: KL to $q^{+}$}, title style={font=\small, yshift=-3pt, text height=1.8ex, text depth=0.3ex},
  error bars/y dir=both, error bars/y explicit, error bars/error bar style={black!60, line width=0.4pt},
  error bars/error mark options={rotate=90, black!60, mark size=1.2pt},
]
\addplot[fill=vanRed, draw=none] coordinates {(ALFWorld,2.91) += (0,0.22) -= (0,0.20) (WebShop,6.31) += (0,0.67) -= (0,0.67) (ScienceWorld,7.66) += (0,0.74) -= (0,0.69)};
\addplot[fill=r2sGreen, draw=none] coordinates {(ALFWorld,1.60) += (0,0.10) -= (0,0.10) (WebShop,3.65) += (0,0.33) -= (0,0.32) (ScienceWorld,4.15) += (0,0.36) -= (0,0.34)};
\end{axis}
\end{tikzpicture}
\hfill
\begin{tikzpicture}
\begin{axis}[
  width=5.1cm, height=2.35cm, scale only axis, ybar, bar width=9pt, bar shift auto, enlarge x limits=0.22,
  symbolic x coords={ALFWorld,WebShop,ScienceWorld}, xtick=data, ymin=0, ymax=70, ytick={0,20,40,60},
  axis lines*=left, axis line style={black!45}, tick style={draw=none}, ytick pos=left,
  ymajorgrids, grid style={densely dotted, black!25}, every tick label/.append style={font=\scriptsize},
  ylabel={Episodes (\%)}, ylabel style={font=\scriptsize, yshift=-4pt},
  title={Student: Out of Turns}, title style={font=\small, yshift=-3pt, text height=1.8ex, text depth=0.3ex},
  error bars/y dir=both, error bars/y explicit, error bars/error bar style={black!60, line width=0.4pt},
  error bars/error mark options={rotate=90, black!60, mark size=1.2pt},
]
\addplot[fill=vanRed, draw=none] coordinates {(ALFWorld,34.18) +- (0,2.65) (WebShop,55.87) +- (0,4.28) (ScienceWorld,33.24) +- (0,2.59)};
\addplot[fill=r2sGreen, draw=none] coordinates {(ALFWorld,18.10) +- (0,2.36) (WebShop,27.53) +- (0,2.66) (ScienceWorld,18.02) +- (0,1.13)};
\end{axis}
\end{tikzpicture}

{\scriptsize \tikz[baseline=-0.4ex]{\fill[vanRed] (0,-0.02) rectangle (0.3,0.16);}\,Vanilla OPD\quad\tikz[baseline=-0.4ex]{\fill[r2sGreen] (0,-0.02) rectangle (0.3,0.16);}\,\method{}}
\caption{\textbf{Repair moves the teacher toward its informed targets, and its student runs out of turns about half as often.} Left: each teacher's top-16 KL to $q^{+}$ on held-out states, with 95\% goal-bootstrap intervals. Right: held-out episodes that end unfinished at the turn budget, mean $\pm$ std over three decodes.}
\label{fig:scorer-effect}
\end{figure}

\textbf{Repair brings the teacher closer to its informed targets, and its student runs out of turns less often.} Scored against the same $q^{+}$ targets over the complete assistant suffix of held-out states, the repaired teacher is closer to them than the unrepaired teacher in every environment, as the left panel of \Cref{fig:scorer-effect} shows. In the evaluation episodes behind \Cref{tab:main}, the student trained on the repaired teacher runs out of turns about half as often as the vanilla OPD student, as the right panel shows.

\begin{table}[t]
\centering
\caption{\textbf{The vanilla student inherits the teacher's endings; the repaired student does not.} Held-out episodes that run out of turns (\%) and turns per episode, seen split, over three decodes.}
\label{tab:timeouts}
\small
\setlength{\tabcolsep}{4pt}
\begin{tabular}{@{}lcccccc@{}}
\toprule
\multirow{2}{*}{Method} & \multicolumn{2}{c}{ALFWorld} & \multicolumn{2}{c}{WebShop} & \multicolumn{2}{c}{ScienceWorld} \\
\cmidrule(lr){2-3}\cmidrule(lr){4-5}\cmidrule(l){6-7}
 & Out of turns & Turns & Out of turns & Turns & Out of turns & Turns \\
\midrule\midrule
Teacher alone & $39.8{\pm}1.5$ & $20.0$ & $56.7{\pm}0.9$ & $11.9$ & $35.7{\pm}1.9$ & $19.7$ \\
Student alone & $55.0{\pm}3.6$ & $22.3$ & $56.7{\pm}1.2$ & $11.9$ & $38.6{\pm}2.1$ & $17.7$ \\
\midrule
Vanilla OPD & \secondscore{$34.3{\pm}4.7$} & \secondscore{$19.0$} & \secondscore{$55.9{\pm}4.3$} & \secondscore{$11.8$} & \secondscore{$31.9{\pm}1.5$} & \secondscore{$18.9$} \\
\textbf{\textcolor{box_green}{\method{} (ours)}} & \bestscore{$19.5{\pm}2.2$} & \bestscore{$14.7$} & \bestscore{$27.5{\pm}2.7$} & \bestscore{$9.0$} & \bestscore{$17.1{\pm}0.7$} & \bestscore{$14.5$} \\
\bottomrule
\end{tabular}
\end{table}

\textbf{The vanilla student inherits the teacher's endings.} \Cref{tab:timeouts} counts, for the same evaluation episodes, how often each policy runs out of turns and how many turns it takes, beside the teacher acting alone and the untrained student. On ALFWorld every failure is a timeout, since the wrapper ends an episode only on success. In every environment the vanilla student takes about the teacher's number of turns and runs out of turns almost as often as the teacher does. The \method{} student finishes in fewer turns and runs out of turns far less often than the teacher that scores it.

\textbf{The saved turns are searching turns.} \Cref{tab:search-turns} splits the turns of the students' training rollouts into searching, that is moving, opening and looking, or searching and paging on WebShop, and acting on the task. The \method{} student spends fewer turns searching than the vanilla OPD student in every environment and about as many on the task. Counting only the episodes each student wins, it still searches less: $8.4$ turns against $10.5$ on ALFWorld, $2.0$ against $3.0$ on WebShop and $7.1$ against $9.1$ on ScienceWorld. On ALFWorld it opens fewer receptacles, on ScienceWorld it looks, reads and waits less, and on WebShop it searches and pages less and buys more. This matches the direction of the informed scores in \Cref{tab:drift}, which fall most on the student's hedging and searching words.

\begin{table}[h]
\centering
\caption{\textbf{The repaired student spends fewer turns searching.} Turns per episode in the last ten training updates, mean$\pm$std over three training runs. \emph{Task} turns act on the goal, such as taking or buying.}
\label{tab:search-turns}
\vspace{2pt}
\small
\setlength{\tabcolsep}{2.5pt}
\begin{tabular}{@{}lccccccccc@{}}
\toprule
\multirow{2}{*}{Method} & \multicolumn{3}{c}{ALFWorld} & \multicolumn{3}{c}{WebShop} & \multicolumn{3}{c}{ScienceWorld} \\
\cmidrule(lr){2-4}\cmidrule(lr){5-7}\cmidrule(l){8-10}
 & Turns & Searching & Task & Turns & Searching & Task & Turns & Searching & Task \\
\midrule\midrule
Vanilla OPD & \secondscore{$20.7{\pm}0.9$} & \secondscore{$15.9{\pm}1.2$} & \secondscore{$4.0{\pm}0.3$} & \secondscore{$11.5{\pm}0.4$} & \secondscore{$7.8{\pm}0.5$} & \secondscore{$3.6{\pm}0.1$} & \secondscore{$18.2{\pm}0.5$} & \secondscore{$10.9{\pm}0.6$} & \secondscore{$6.9{\pm}0.7$} \\
\method{} & \bestscore{$15.6{\pm}0.5$} & \bestscore{$11.4{\pm}0.6$} & \bestscore{$3.7{\pm}0.3$} & \bestscore{$8.8{\pm}0.2$} & \bestscore{$4.1{\pm}0.1$} & \bestscore{$4.7{\pm}0.1$} & \bestscore{$14.6{\pm}0.3$} & \bestscore{$8.2{\pm}0.6$} & \bestscore{$6.1{\pm}0.3$} \\
\bottomrule
\end{tabular}
\end{table}

\section{Conclusion}
\label{sec:conclusion}

The teacher's takeovers can still finish from the states where the student departs most from it,
yet its plain scores let the student's delay pass; only a finishing takeover in context turns them against it.
\method{} transfers information from selected teacher continuations into the
scores used by ordinary OPD. Its repair supervises every assistant
position from the divergence turn onward with takeover-informed targets,
then freezes the repaired teacher as the student's sole scorer. Under a shared protocol,
\method{} outperforms three state-of-the-art agentic-OPD methods in all twelve model--environment cells,
and it also leads when the teacher is the student's own base model.

\section*{Limitations}
\textbf{Compute.} Before distillation, \method{} spends the untrained student's episodes, $N{\times}K=19{,}200$ takeover episodes, the selection step and one teacher fine-tune. Every method then trains for 150 steps; Guided-OPD and FutureBridge also spend teacher generation inside training. \textbf{One model family.} Evidence comes from text-agent benchmarks and Qwen3 and Qwen3.5 models. \textbf{Replayable environments.} Takeovers start from the student's own states, so \method{} needs an environment that can be replayed to a given state; all three benchmarks here can.

\bibliography{references}
\bibliographystyle{iclr2027_conference}

\appendix
\crefalias{section}{appendix}\crefalias{subsection}{appendix}
\crefname{appendix}{Appendix}{Appendices}\Crefname{appendix}{Appendix}{Appendices}

\clearpage
\section{Extended Related Work}
\label{app:relwork}

\subsection{Detailed Comparison with the Closest Methods}
\label{app:compare}

\begin{table}[t]
\centering
\caption{\textbf{Among these methods, only \method{} repairs an external teacher without task reward and refreezes it for distillation.} RQ-free: no task-reward query for distillation supervision. The two right columns concern an external scorer, not an evolving self-teacher. $^{\dagger}$FutureBridge uses a success-selected reference pool.}
\label{tab:positioning}
\footnotesize
\setlength{\tabcolsep}{4pt}
\begin{tabular}{@{}>{\raggedright\arraybackslash}p{5.8cm}ccc@{}}
\toprule
\textbf{Method} & \textbf{RQ-free} & \shortstack{\textbf{Repairs external}\\\textbf{teacher}} & \shortstack{\textbf{Freeze, then}\\\textbf{distill}} \\
\midrule\midrule
TCOD-F2B \citep{tcod2026} & \textcolor{box_green}{\cmark} & \textcolor{gray}{\xmark} & \textcolor{gray}{\xmark} \\
Guided-OPD \citep{guidedopd2026} & \textcolor{box_green}{\cmark} & \textcolor{gray}{\xmark} & \textcolor{gray}{\xmark} \\
FutureBridge$^{\dagger}$ \citep{futurebridge2026} & \textcolor{box_green}{\cmark} & \textcolor{gray}{\xmark} & \textcolor{gray}{\xmark} \\
ReOPD \citep{reopd2026} & \textcolor{box_green}{\cmark} & \textcolor{gray}{\xmark} & \textcolor{gray}{\xmark} \\
SDFT / OPCD \citep{sdft2026,opcd2026} & \textcolor{box_green}{\cmark} & \textcolor{gray}{\xmark} & \textcolor{gray}{\xmark} \\
CANON \citep{canon2026} & \textcolor{box_green}{\cmark} & \textcolor{gray}{\xmark} & \textcolor{gray}{\xmark} \\
SDAR \citep{sdar2026} & \textcolor{gray}{\xmark} & \textcolor{gray}{\xmark} & \textcolor{gray}{\xmark} \\
Distillation Traps and Guards \citep{trapsguards2026} & \textcolor{gray}{\xmark} & \textcolor{box_green}{\cmark} & \textcolor{box_green}{\cmark} \\
\midrule
\textbf{\method{} (ours)} & \textcolor{box_green}{\cmark} & \textcolor{box_green}{\cmark} & \textcolor{box_green}{\cmark} \\
\bottomrule
\end{tabular}
\end{table}

\Cref{tab:positioning} places the compared methods by the three properties in its columns; updating an EMA self-teacher during actor training is different from this separate scorer-repair stage.

\textbf{Student-side curricula.} TCOD-F2B~\citep{tcod2026} gradually lengthens the student's training episodes. Guided-OPD~\citep{guidedopd2026} assigns a decreasing share of turns to the teacher, combining cross-entropy on teacher tokens with distillation on student tokens. ReOPD~\citep{reopd2026} replays pre-collected teacher prefixes. All three leave the external scorer unchanged; a repaired teacher can supply scores under such curricula as well as under ordinary OPD.

\textbf{FutureBridge and teacher continuations.} FutureBridge~\citep{futurebridge2026} is our closest experimental comparator. It continues from high-disagreement student states and validates a bridge with short student rollouts; accepted teacher text enters the student's training buffer, alongside a curriculum over success-selected reference prefixes. Its online acceptance rule reads no task reward, although the reference pool uses success selection. \method{} instead uses a selected takeover as privileged context for teacher repair: the teacher learns to reproduce its informed scores without that context and is then frozen, and the student retains ordinary OPD.

\textbf{Privileged context and consensus.} SDFT~\citep{sdft2026} and Self-Distilled Reasoner~\citep{opsd2026} train an acting model from a demonstration-conditioned self-teacher. OPCD~\citep{opcd2026} also supports an external context-conditioned teacher, but transfers the context into the student directly. CANON~\citep{canon2026} supplies a label-free reference by selecting the solution with the majority answer. Our consensus selection follows this principle, applied to an external teacher's continuations from student states; the model being fitted becomes the scorer of a subsequent distillation stage.

\textbf{Feedback and reward-based learning.} SDPO~\citep{sdpo2026} turns rich environment feedback into self-distillation targets, and in scalar-feedback settings uses successful attempts as feedback. SDAR~\citep{sdar2026} is a multi-turn hybrid that keeps reward-based RL as its primary objective and adds gated self-distillation. We exclude this additional student reward signal from the experimental comparison; \method{} could supply the distillation scorer of such a hybrid.

\textbf{External teacher calibration.} Distillation Traps and Guards~\citep{trapsguards2026} also changes the teacher before distillation, by reinforcement fine-tuning with utility and calibration rewards and a KL anchor; \method{} instead fits takeover-informed soft targets on student histories through the cross-entropy of \Cref{eq:r2s}, without task reward.

\subsection{Further Related Work}
\label{app:related}

\textbf{OPD at scale.} MiMo-V2-Flash integrates domain-specialized teachers with dense token-level supervision~\citep{mimo2026}, and Nemotron-Cascade 2 uses multi-domain OPD between Cascade RL stages to recover capabilities from the strongest intermediate domain checkpoints~\citep{nemotroncascade22026}. Our question concerns the quality of the next-token supervision a given teacher supplies on multi-turn student histories.

\textbf{Other ways to adapt a model.} OPD is one of several routes from a pretrained model to a task. Fine-tuning adapts the weights on fixed data, through low-rank adapters, knowledge-anchored low-resource adaptation or prompt-based continual learning~\citep{zhao2026mora,zhao2025kalahash,zhao2024reconstruct,zou2025learning}; RL adapts them on task reward, as in ParallelSearch for search agents~\citep{zhao2025parallelsearch}; in-context learning leaves the weights fixed and adapts through examples, including self-mined errors~\citep{zhao2026value}; and task-specific models are built for one domain~\citep{hao2025safemap}. Agentic OPD differs from all four in its signal: a teacher's next-token distribution on the student's own trajectories, scored while the student acts.

\textbf{Local supervision and privileged information.} Beyond the compatibility and privileged-context findings of \citet{li2026bestval} and \citet{rethinkingopsd2026} discussed in \Cref{sec:related}, Multi-Rollout OPD~\citep{mropd2026} enriches the teacher's context with peer attempts labelled by success or failure. Our references are teacher takeovers from the student's own state, selected without task reward and internalized into an external scorer, which we assess through target fidelity and the completion behavior of its distilled students.

\section{Analysis Details}
\label{app:analysis}
\label{app:delay}

\begin{table}[t]
\centering
\caption{\textbf{The states and scores behind \Cref{tab:drift}.} \emph{All}: the untrained student's highest-disagreement states with a replayable takeover; \emph{Turn $\geq 5$}: those from turn five on, which \Cref{tab:drift} uses; \emph{Lost}: of these, the states the student went on to lose. \emph{Scored}: states where at least one takeover finishes, and the student's turns through the mined one, which the drop columns read. \emph{Against other}: a class's mean change in log-probability over that of the rest of the thinking, with a 95\% interval over states; \emph{Drop}: as in \Cref{tab:drift}, for the student's committing words and its action tokens.}
\label{tab:drift-detail}
\footnotesize
\setlength{\tabcolsep}{3.5pt}
\begin{tabular}{@{}lccccccccc@{}}
\toprule
& \multicolumn{3}{c}{States} & \multicolumn{2}{c}{Scored} & \multicolumn{2}{c}{Against other} & \multicolumn{2}{c}{Drop (\%)} \\
\cmidrule(lr){2-4}\cmidrule(lr){5-6}\cmidrule(lr){7-8}\cmidrule(l){9-10}
Environment & All & Turn $\geq 5$ & Lost & States & Turns & Hedging & Searching & Committing & Action \\
\midrule\midrule
ALFWorld & 2,399 & 814 & 489 & 458 & 4,957 & $2.4$ $[2.1, 2.8]$ & $1.8$ $[1.4, 2.2]$ & $6.1$ & $0.1$ \\
WebShop & 2,398 & 689 & 647 & 107 & 906 & $2.0$ $[1.3, 2.9]$ & $1.9$ $[0.9, 3.0]$ & $6.8$ & $0.5$ \\
ScienceWorld & 2,205 & 682 & 555 & 403 & 4,085 & $2.7$ $[2.4, 3.1]$ & $3.5$ $[3.0, 4.1]$ & $11.3$ & $0.3$ \\
\bottomrule
\end{tabular}
\end{table}

\textbf{The table of \Cref{sec:takeover}.} \Cref{tab:drift} uses the untrained student's highest-disagreement states from turn five on, where a takeover is no longer close to the teacher acting alone from the start; \Cref{tab:drift-detail} counts them. Success is the environment's own outcome. The drop columns keep the states where at least one takeover finishes and read the student's turns up to and including the mined one. The teacher scores the student's own tokens twice, plain and with that takeover in context, as described in \Cref{app:r2s}; a drop entry is $100\,(1-e^{\bar d})$, where $\bar d$ is the mean change in log-probability over the student's tokens of that kind. Hedging words are the thinking tokens \emph{maybe, might, likely, probably, perhaps, could, possibly, seem(s), guess, assume, unsure, potentially, whether, if, unless, unclear, unknown, hopefully, presumably}; searching words are \emph{look, search, check, examine, explore, find, locate, scan, browse} with their inflections, and \emph{inventory}; \emph{other} is the rest of the thinking. \Cref{tab:drift-detail} also gives, with 95\% intervals over states, how much more each class falls than the rest of the thinking, and the drop on the student's words of committing (\emph{take, put, use, buy, now, done}, \ldots), which fall less than its hedging words and more than the rest, and on its action tokens, which barely move.

\begin{table}[t]
\centering
\caption{\textbf{The teacher's takeovers finish from the student's states in every cell.} Success (\%) of the untrained student and of the teacher's $K{=}8$ takeovers over all mined states, for every pair of \Cref{tab:main}.}
\label{tab:takeover-all}
\small
\setlength{\tabcolsep}{4.5pt}
\begin{tabular}{@{}lcccccc@{}}
\toprule
\multirow{2}{*}{Teacher $\to$ student} & \multicolumn{2}{c}{ALFWorld} & \multicolumn{2}{c}{WebShop} & \multicolumn{2}{c}{ScienceWorld} \\
\cmidrule(lr){2-3}\cmidrule(lr){4-5}\cmidrule(l){6-7}
 & Student & Takeover & Student & Takeover & Student & Takeover \\
\midrule\midrule
Qwen3.5-9B$\to$2B & $11.9$ & $54.6$ & $5.0$ & $21.8$ & $1.2$ & $31.2$ \\
Qwen3.5-9B$\to$4B & $51.8$ & $56.7$ & $19.6$ & $22.0$ & $22.4$ & $35.6$ \\
Qwen3-30B-A3B$\to$4B & $33.3$ & $41.8$ & $10.4$ & $14.0$ & $10.2$ & $36.6$ \\
Qwen3-30B-A3B$\to$8B & $45.5$ & $42.2$ & $20.1$ & $31.7$ & $21.6$ & $38.8$ \\
\bottomrule
\end{tabular}
\end{table}

\textbf{The takeovers of every pair.} \Cref{tab:takeover-all} repeats the student-versus-takeover comparison of \Cref{tab:drift} for every cell of \Cref{tab:main}, over all mined states rather than those from turn five on; both entries of its Qwen3.5-9B$\to$4B row are therefore above the matching entries of \Cref{tab:drift}, since the earlier states are easier for the student and for the teacher alike. The $K{=}8$ takeovers beat the student's own outcome in eleven of twelve cells; the exception is the Qwen3-8B student on ALFWorld, which already beats its teacher acting alone and which \method{} still lifts ten points above vanilla OPD. On WebShop the 30B-A3B teacher, which seldom finishes from the first turn, finishes $14\%$ and $32\%$ of its takeovers from the students' states.

\textbf{How the episodes end.} The out-of-turn shares that \Cref{sec:takeover} cites are those of \Cref{fig:scorer-effect} and \Cref{tab:timeouts}: held-out evaluation episodes that reach the turn budget unfinished, for the 9B-taught 4B students of \Cref{tab:main} at their final checkpoint, in the three decodes behind their success there; \Cref{fig:scorer-effect} averages the seen and unseen splits, \Cref{tab:timeouts} reports the seen split.

\textbf{The \method{} student spends fewer turns searching.} \Cref{tab:search-turns} reads the students' own training rollouts in the last ten updates of each run. \emph{Searching} counts the turns spent moving, opening and looking (ALFWorld: go to, open, close, look, examine, inventory; ScienceWorld: teleport, look, read, wait, open, close; WebShop: search, next or previous page, back to search); \emph{task} counts the turns spent acting on the goal (ALFWorld: take, move, put, clean, heat, cool, use; ScienceWorld: focus, pick up, move, pour, use, activate, connect and the like; WebShop: opening an item, choosing an option, reading its description, buying). The win-only counts of \Cref{sec:exp-longhorizon} are averaged over the three runs in the same way.

\section{\method{} Implementation Details}
\label{app:r2s}

\textbf{Untrained-student episodes.} Before training, the untrained student plays the training-split tasks under the training prompt at $T{=}1.0$, for the episode budget of a training run (150 steps); evaluation tasks are discarded before scoring.

\textbf{Mining score.} In $\Delta_t$ of \Cref{eq:divscore}, $\student$ is the untrained student, and each term is the sampled per-token estimate of the reverse KL of \Cref{eq:opd}, so a single token's term can take either sign. Action tokens stay out of $\Delta_t$ because the thinking before them decides the action; with a finishing takeover in context, the teacher's scores of the student's action tokens barely move, as \Cref{tab:drift-detail} shows.

\textbf{Filters and mining.} A turn is eligible when its thinking span has at least three tokens and when it ends in a tool call. Within each episode we keep the eligible turn with the largest $\Delta_t$ (one state per episode), cap three states per goal, and keep the top $N{=}2{,}400$ states by $\Delta_t$. A state is dropped when its prompt cannot be re-tokenized identically, when its privileged stream exceeds $32{,}768$ tokens, or when it has no privileged position.

\textbf{Takeovers.} From each restored state the teacher attempts the episode $K{=}8$ times independently, sampled at $T{=}1.0$ with reasoning mode off, up to the environment's turn budget. \Cref{tab:takeover-all} gives the takeover success of every cell of \Cref{tab:main}; between $2{,}159$ and $2{,}400$ of a cell's mined states have at least one replayable takeover, and at least one of the eight takeovers finishes in $28$--$83\%$ of them, depending on the cell.

\textbf{Selection.} We use two reward-free checks. The consensus vote groups a state's takeovers by how they end and keeps the state when the largest group holds at least $m$ of them; the LLM judge, Qwen3.5-9B with reasoning mode on, reads each takeover and keeps the state when it approves at least $m{=}1$. For \Cref{tab:main}, the vote uses $m{=}2$ on ALFWorld, over the takeovers that end within the turn budget, and $m{=}1$ on ScienceWorld, over those that end with the environment's completion message; on WebShop the judge checks each purchase against the instruction, for both teachers, and the approved takeovers are grouped by the item bought. The shortest takeover of the largest group is selected. In \Cref{tab:selection}, the consensus row votes with $m{=}2$ everywhere, and on ALFWorld and ScienceWorld the judge row reads the whole episode but never the reward or the completion message, and takes the shortest approved takeover. Only the ground-truth row of \Cref{tab:selection} reads the task reward.

\textbf{Privileged targets and plain scores.} The frozen teacher $T_0$ scores the student's unmodified token stream through the end of the episode. In the privileged pass the selected takeover is inserted as one extra user message immediately before the student's first turn. The message is headed ``Reference walkthrough'' and lists the teacher's thinking and action for each takeover step, without the replayed prefix and without environment observations; the header presents it as a successful expert completion of the same task. Every position of the student stream shifts by one constant offset, so the two passes align position by position. The plain pass scores the same stream without the message. Both passes cache per-position top-16 (token, log-probability) lists. Every assistant position from the mined turn, inclusive, to the student's final assistant turn receives $q^+$, including assistant turns without a tool call; earlier history remains causal context, and prompts and environment observations are not targets. \method{} trains only on the privileged targets; the plain pass is the reference of the drop columns of \Cref{tab:drift}.

\textbf{Loss and fine-tuning.} The loss at each labeled position is the cross-entropy between the target $q^+$, kept on its top-16 tokens and renormalized, and the model's full-vocabulary log-softmax. The targets are computed once from the frozen $T_0$ and are not recomputed as $T_\phi$ trains. As in \Cref{eq:r2s}, we average all selected assistant positions within each state, then average states. The teacher is initialized from $T_0$ and fine-tuned full-parameter for one epoch at learning rate $2\times10^{-6}$, with AdamW (betas $0.9/0.95$, no weight decay), 10 warmup steps followed by cosine decay to $0.1\times$, gradient clip 1.0, batches of 8 labeled states and bf16; 10\% of goals form a goal-disjoint held-out split. For Qwen3.5-9B the vision tower stays frozen, so full-parameter means the text decoder. The label-SFT control of \Cref{tab:abl-components} uses the same recipe. The repaired teacher is then refrozen and scores otherwise ordinary OPD in place of $T_0$, with the sampled estimator of \Cref{eq:opd-sampled}; the original teacher takes no part in the student's training.

\textbf{Self-distillation.} In \Cref{tab:self,tab:self-full}, teacher and untrained student are the same Qwen3.5-4B checkpoint, so $\Delta_t$ of \Cref{eq:divscore} is zero up to numerical noise and cannot rank states. The self pipeline instead draws one eligible turn per episode of the untrained student at random, under the same goal cap and budget, and the 4B itself runs the takeovers and the selection, so no larger model enters the pipeline. Repair and distillation follow the settings above. The vanilla control, TCOD-F2B, Guided-OPD and FutureBridge learn from the unrepaired 4B teacher; the Student alone column is the untrained 4B of \Cref{tab:main,tab:main-full}.

\section{Experimental Protocol}
\label{app:protocol}

\textbf{Environments.} WebShop~\citep{webshop2022} reports success on the 500 canonical evaluation goals and on a disjoint 256-goal pool. ALFWorld~\citep{alfworld2021} reports seen and unseen success on its held-out games under a TCOD-aligned full-history protocol; our wrapper ends an episode only on success. ScienceWorld~\citep{scienceworld2022} follows the TCOD splits: the first half of each of the 17 training task types' variations trains, 140 variations sampled from the second half form the seen evaluation, and the first 134 rows of the frozen 1{,}308-task test set of unseen types form the unseen evaluation; the metric is the simulator's completion flag, and the simulator and the action and object lists shown to the agent are TCOD's.

\textbf{Models.} The models are those of \Cref{sec:exp-setup}. Model and tokenizer revisions, prompts, environment wrappers and checkpoint provenance are in the code release, tag \texttt{r2s} of \href{https://github.com/Tree-Shu-Zhao/miles-agent}{\nolinkurl{github.com/Tree-Shu-Zhao/miles-agent}}; its 24 checkpoints are on the Hugging Face Hub.

\textbf{Prompt protocol.} Every arm and anchor of \Cref{tab:main} uses one prompt per environment, rendered in each model's native tool-call template; the prompt asks for one to three sentences of plain-text deliberation before exactly one tool call per turn, and that deliberation stays part of the generated response.

\textbf{Objective.} The vanilla OPD, \method{} and TCOD-F2B rows of \Cref{tab:main} minimize the reverse KL of \Cref{eq:opd} with its sampled estimator, which gives each student token the advantage~\citep{mimo2026,li2026bestval}
\begin{equation}
\label{eq:opd-sampled}
A_{t,i}=-\beta\!\left[\log \student(y_{t,i}\mid h_{t,i})-\log \teacher(y_{t,i}\mid h_{t,i})\right],
\qquad \tau\sim\student,
\end{equation}
with coefficient $\beta{=}1$, observations loss-masked and updates restricted to student tokens. Guided-OPD selects the teacher or the student to act at each turn and trains the student with cross-entropy on teacher-generated tokens and the same sampled reverse KL on student-generated tokens; FutureBridge applies the same sampled reverse KL to episodes that carry an accepted teacher bridge. Task reward is logged and zeroed for every distillation arm.

\textbf{Training.} Every distillation run takes 150 optimization steps and evaluates every 25 steps; the 128 WebShop episodes per step are $16$ goals $\times$ $8$ samples. Adam uses betas $(0.9, 0.98)$, weight decay 0.1 and gradient clip 1.0, and the PPO ratio clip is 0.2. TCOD-F2B, Guided-OPD and FutureBridge share this protocol and keep their own method settings.

\textbf{Evaluation.} Student-alone anchors are three decodes of the untrained student under the same prompt. An evaluation in which an episode returned no outcome after an infrastructure failure is repeated at unchanged weights and decoding seed.

\textbf{Teacher acting alone.} The teacher itself is evaluated as the acting policy at $T{=}0.4$; its split means are the teacher-alone anchors of \Cref{tab:main}, and \Cref{tab:main-full} gives them per split.

\section{Every Split of the Main Results}
\label{app:results}
\label{app:splits}

\textbf{The main text averages the splits.} Each cell of \Cref{tab:main} and \Cref{tab:self} is, per decoding seed, the mean of the seen and unseen values of the same evaluation (ALFWorld, ScienceWorld) or the success rate on the 500 canonical WebShop goals; the mean and std are then taken over the three decoding seeds. \Cref{tab:main-full} and \Cref{tab:self-full} give every split, and for WebShop also the disjoint 256-goal pool. \method{} exceeds every baseline in mean on all $24$ splits.

\begin{table}[t]
\centering
\caption{\textbf{Main results by split.} The cells of \Cref{tab:main} before averaging: ALFWorld and ScienceWorld seen\,/\,unseen, WebShop the 500 canonical goals\,/\,the disjoint 256-goal pool; protocol as in \Cref{tab:main}. Bold: the best of the five trained arms where its gap to the runner-up exceeds the larger decode std; ties are unmarked.}
\label{tab:main-full}
\scriptsize
\setlength{\tabcolsep}{3.5pt}
\begin{tabular}{@{}llcccccc@{}}
\toprule
Student & Split & Student alone & Vanilla OPD & TCOD-F2B & Guided-OPD & FutureBridge & \method{} (ours) \\
\midrule\midrule
\multicolumn{8}{@{}l}{\textbf{Panel A: Qwen3.5-9B $\to$ Qwen3.5-2B\,/\,4B, three environments}} \\
\multicolumn{4}{@{}l}{ALFWorld (success, \%)} & \multicolumn{4}{r@{}}{\textit{teacher alone: seen $60.2{\pm}1.5$, unseen $65.9{\pm}3.0$}} \\
\multirow{2}{*}{Qwen3.5-2B} & seen & $19.8{\pm}1.5$ & $55.0{\pm}2.6$ & $59.0{\pm}2.2$ & $60.2{\pm}4.8$ & $54.0{\pm}2.9$ & $\mathbf{76.9{\pm}3.2}$ \\
 & unseen & $10.4{\pm}3.7$ & $60.9{\pm}2.3$ & $66.4{\pm}2.0$ & $66.9{\pm}1.1$ & $59.5{\pm}4.3$ & $\mathbf{78.9{\pm}3.0}$ \\
\addlinespace[1.5pt]
\multirow{2}{*}{Qwen3.5-4B} & seen & $45.0{\pm}3.6$ & $65.7{\pm}4.7$ & $59.3{\pm}2.6$ & $60.2{\pm}4.4$ & $58.1{\pm}2.9$ & $\mathbf{80.5{\pm}2.2}$ \\
 & unseen & $41.8{\pm}3.9$ & $65.9{\pm}1.1$ & $60.0{\pm}1.6$ & $62.9{\pm}1.1$ & $58.7{\pm}4.5$ & $\mathbf{83.3{\pm}3.1}$ \\
\addlinespace[3pt]
\multicolumn{4}{@{}l}{WebShop (success, \%; 500 canonical goals\,/\,256-goal pool)} & \multicolumn{4}{r@{}}{\textit{teacher alone: canonical $33.5{\pm}0.3$, pool $28.6{\pm}2.4$}} \\
\multirow{2}{*}{Qwen3.5-2B} & canonical & $17.6{\pm}1.1$ & $35.7{\pm}1.6$ & $33.1{\pm}0.5$ & $35.6{\pm}0.6$ & $36.3{\pm}1.0$ & $\mathbf{57.9{\pm}1.6}$ \\
 & pool & $16.0{\pm}4.1$ & $33.5{\pm}0.2$ & $29.9{\pm}1.6$ & $32.0{\pm}1.7$ & $31.9{\pm}2.5$ & $\mathbf{52.2{\pm}1.8}$ \\
\addlinespace[1.5pt]
\multirow{2}{*}{Qwen3.5-4B} & canonical & $33.7{\pm}0.9$ & $34.6{\pm}2.9$ & $35.1{\pm}0.9$ & $35.5{\pm}1.3$ & $33.9{\pm}1.7$ & $\mathbf{55.9{\pm}1.4}$ \\
 & pool & $31.5{\pm}1.6$ & $32.8{\pm}1.8$ & $32.7{\pm}0.6$ & $31.4{\pm}1.9$ & $30.5{\pm}1.4$ & $\mathbf{54.7{\pm}1.7}$ \\
\addlinespace[3pt]
\multicolumn{4}{@{}l}{ScienceWorld (success, \%)} & \multicolumn{4}{r@{}}{\textit{teacher alone: seen $33.6{\pm}3.6$, unseen $30.3{\pm}0.4$}} \\
\multirow{2}{*}{Qwen3.5-2B} & seen & $2.4{\pm}1.1$ & $25.5{\pm}1.1$ & $34.0{\pm}2.5$ & $31.2{\pm}4.9$ & $18.1{\pm}5.5$ & $\mathbf{41.7{\pm}1.5}$ \\
 & unseen & $3.7{\pm}1.3$ & $21.9{\pm}4.1$ & $21.9{\pm}3.4$ & $20.6{\pm}2.6$ & $19.9{\pm}1.7$ & $\mathbf{34.1{\pm}0.4}$ \\
\addlinespace[1.5pt]
\multirow{2}{*}{Qwen3.5-4B} & seen & $26.7{\pm}1.8$ & $38.8{\pm}2.2$ & $32.1{\pm}0.7$ & $35.0{\pm}3.7$ & $32.4{\pm}0.8$ & $\mathbf{52.6{\pm}3.5}$ \\
 & unseen & $29.4{\pm}1.1$ & $30.3{\pm}5.0$ & $29.1{\pm}1.3$ & $29.4{\pm}1.6$ & $29.1{\pm}2.0$ & $\mathbf{43.0{\pm}3.0}$ \\
\midrule\midrule
\multicolumn{8}{@{}l}{\textbf{Panel B: Qwen3-30B-A3B-Instruct-2507 $\to$ Qwen3-4B\,/\,8B, three environments}} \\
\multicolumn{4}{@{}l}{ALFWorld (success, \%)} & \multicolumn{4}{r@{}}{\textit{teacher alone: seen $43.1{\pm}2.2$, unseen $41.5{\pm}2.6$}} \\
\multirow{2}{*}{Qwen3-4B} & seen & $27.4{\pm}0.4$ & $41.9{\pm}0.8$ & $41.7{\pm}1.1$ & $42.9{\pm}1.4$ & $44.5{\pm}0.4$ & $\mathbf{54.8{\pm}1.1}$ \\
 & unseen & $33.3{\pm}4.2$ & $40.5{\pm}2.8$ & $43.5{\pm}1.1$ & $41.0{\pm}1.3$ & $39.1{\pm}5.6$ & $\mathbf{54.2{\pm}1.9}$ \\
\addlinespace[1.5pt]
\multirow{2}{*}{Qwen3-8B} & seen & $45.2{\pm}1.1$ & $46.0{\pm}0.4$ & $48.6{\pm}1.9$ & $47.1{\pm}1.9$ & $45.2{\pm}2.2$ & $\mathbf{56.4{\pm}0.7}$ \\
 & unseen & $48.5{\pm}0.7$ & $43.0{\pm}4.8$ & $49.0{\pm}2.4$ & $46.8{\pm}5.1$ & $46.5{\pm}2.8$ & $\mathbf{53.0{\pm}1.3}$ \\
\addlinespace[3pt]
\multicolumn{4}{@{}l}{WebShop (success, \%; 500 canonical goals\,/\,256-goal pool)} & \multicolumn{4}{r@{}}{\textit{teacher alone: canonical $2.9{\pm}0.3$, pool $3.0{\pm}1.1$}} \\
\multirow{2}{*}{Qwen3-4B} & canonical & $15.5{\pm}1.2$ & $24.7{\pm}1.1$ & $22.8{\pm}0.7$ & $23.3{\pm}1.6$ & $16.7{\pm}0.9$ & $\mathbf{28.5{\pm}0.6}$ \\
 & pool & $13.8{\pm}0.5$ & $21.6{\pm}1.5$ & $21.1{\pm}2.0$ & $19.1{\pm}0.4$ & $14.5{\pm}1.4$ & $24.6{\pm}3.5$ \\
\addlinespace[1.5pt]
\multirow{2}{*}{Qwen3-8B} & canonical & $23.6{\pm}1.2$ & $20.3{\pm}0.9$ & $23.4{\pm}0.7$ & $22.3{\pm}0.8$ & $20.7{\pm}0.7$ & $\mathbf{30.4{\pm}1.2}$ \\
 & pool & $23.7{\pm}2.1$ & $19.7{\pm}0.8$ & $19.8{\pm}1.6$ & $18.0{\pm}2.3$ & $18.0{\pm}3.1$ & $\mathbf{27.9{\pm}3.2}$ \\
\addlinespace[3pt]
\multicolumn{4}{@{}l}{ScienceWorld (success, \%)} & \multicolumn{4}{r@{}}{\textit{teacher alone: seen $29.5{\pm}3.2$, unseen $36.3{\pm}3.5$}} \\
\multirow{2}{*}{Qwen3-4B} & seen & $6.9{\pm}2.7$ & $35.2{\pm}4.3$ & $34.3{\pm}2.5$ & $32.6{\pm}4.2$ & $33.1{\pm}1.8$ & $\mathbf{48.3{\pm}1.1}$ \\
 & unseen & $11.2{\pm}0.0$ & $40.8{\pm}2.4$ & $42.5{\pm}0.7$ & $36.8{\pm}2.8$ & $39.1{\pm}2.4$ & $43.3{\pm}2.0$ \\
\addlinespace[1.5pt]
\multirow{2}{*}{Qwen3-8B} & seen & $18.6{\pm}3.6$ & $37.6{\pm}2.1$ & $32.1{\pm}2.9$ & $31.7{\pm}3.7$ & $37.4{\pm}3.4$ & $\mathbf{56.2{\pm}0.4}$ \\
 & unseen & $29.4{\pm}5.2$ & $36.1{\pm}2.4$ & $37.6{\pm}2.4$ & $34.3{\pm}0.7$ & $36.8{\pm}3.4$ & $\mathbf{44.8{\pm}2.6}$ \\
\bottomrule
\end{tabular}
\end{table}

\begin{table}[t]
\centering
\caption{\textbf{Self-distillation, every split.} The cells of \Cref{tab:self} by split (ALFWorld and ScienceWorld seen\,/\,unseen; WebShop canonical goals); protocol as in \Cref{tab:self}. Bold: the best of the five trained arms where its gap to the runner-up exceeds the larger decode std.}
\label{tab:self-full}
\scriptsize
\setlength{\tabcolsep}{3.5pt}
\begin{tabular}{@{}llcccccc@{}}
\toprule
Environment & Split & Student alone & Vanilla OPD & TCOD-F2B & Guided-OPD & FutureBridge & \method{} (ours) \\
\midrule\midrule
\multirow{2}{*}{ALFWorld} & seen & $45.0{\pm}3.6$ & $53.1{\pm}2.3$ & $50.5{\pm}3.2$ & $50.0{\pm}4.5$ & $56.4{\pm}3.8$ & $\mathbf{69.5{\pm}2.3}$ \\
 & unseen & $41.8{\pm}3.9$ & $46.8{\pm}4.1$ & $49.0{\pm}4.1$ & $47.0{\pm}3.4$ & $44.8{\pm}0.0$ & $\mathbf{75.9{\pm}1.6}$ \\
\addlinespace[3pt]
WebShop & canonical & $33.7{\pm}0.9$ & $33.3{\pm}2.2$ & $31.3{\pm}0.9$ & $34.0{\pm}1.8$ & $31.1{\pm}1.5$ & $\mathbf{58.1{\pm}2.5}$ \\
\addlinespace[3pt]
\multirow{2}{*}{ScienceWorld} & seen & $26.7{\pm}1.8$ & $25.5{\pm}1.8$ & $25.7{\pm}2.1$ & $30.0{\pm}0.0$ & $28.1{\pm}3.9$ & $\mathbf{45.2{\pm}5.7}$ \\
 & unseen & $29.4{\pm}1.1$ & $30.8{\pm}4.6$ & $28.6{\pm}1.1$ & $28.6{\pm}3.5$ & $30.3{\pm}2.8$ & $\mathbf{39.6{\pm}2.7}$ \\
\bottomrule
\end{tabular}
\end{table}

\section{Training Curves}
\label{app:curves}
\label{app:curves-all}

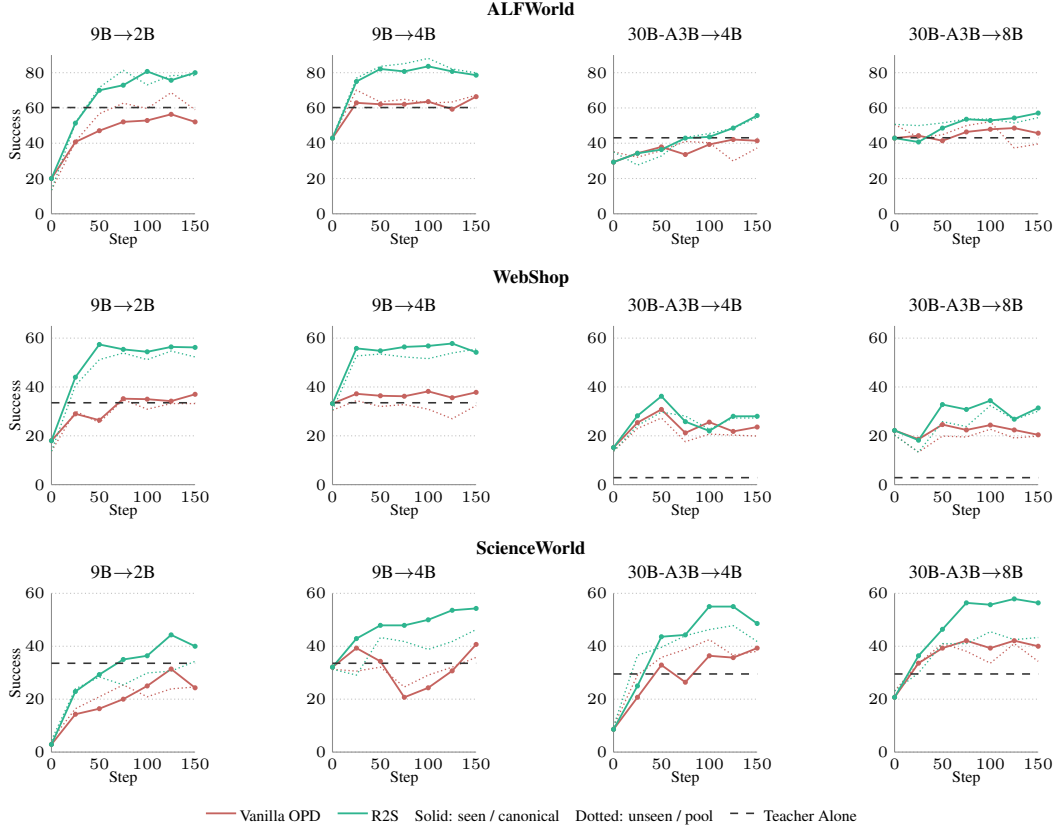
\begin{figure}[t]
\centering
\definecolor{vanRed}{HTML}{C5635F}
\definecolor{r2sGreen}{HTML}{2EB58F}
\scriptsize
\makebox[\linewidth][c]{\textbf{ALFWorld}}\\[1pt]
\begin{tikzpicture}
\begin{axis}[
  width=1.9cm, height=2.1cm, scale only axis,
  xmin=0, xmax=150, ymin=0, ymax=90, xtick={0,50,100,150}, ylabel={Success}, xlabel={Step},
  axis lines*=left, axis line style={black!45}, tick style={draw=none}, ytick pos=left,
  ymajorgrids, grid style={densely dotted, black!25}, every tick label/.append style={font=\tiny},
  ylabel style={font=\tiny, yshift=-3pt}, xlabel style={font=\tiny, yshift=4pt}, title style={font=\scriptsize, yshift=-3pt},
  title={9B$\to$2B},
]
\addplot[vanRed, line width=0.7pt, mark=*, mark size=0.6pt] coordinates {(0,20.0) (25,40.7) (50,47.1) (75,52.1) (100,52.9) (125,56.4) (150,52.1)};
\addplot[vanRed, densely dotted, line width=0.5pt] coordinates {(0,13.4) (25,41.0) (50,56.7) (75,62.7) (100,59.7) (125,68.7) (150,59.0)};
\addplot[r2sGreen, line width=0.7pt, mark=*, mark size=0.6pt] coordinates {(0,20.0) (25,51.4) (50,70.0) (75,72.9) (100,80.7) (125,75.7) (150,80.0)};
\addplot[r2sGreen, densely dotted, line width=0.5pt] coordinates {(0,13.4) (25,51.5) (50,71.6) (75,81.3) (100,73.1) (125,78.4) (150,78.4)};
\draw[dashed, black!80, line width=0.6pt] (rel axis cs:0,0.6693) -- (rel axis cs:1,0.6693);
\end{axis}
\end{tikzpicture}
\hfill
\begin{tikzpicture}
\begin{axis}[
  width=1.9cm, height=2.1cm, scale only axis,
  xmin=0, xmax=150, ymin=0, ymax=90, xtick={0,50,100,150},  xlabel={Step},
  axis lines*=left, axis line style={black!45}, tick style={draw=none}, ytick pos=left,
  ymajorgrids, grid style={densely dotted, black!25}, every tick label/.append style={font=\tiny},
  ylabel style={font=\tiny, yshift=-3pt}, xlabel style={font=\tiny, yshift=4pt}, title style={font=\scriptsize, yshift=-3pt},
  title={9B$\to$4B},
]
\addplot[vanRed, line width=0.7pt, mark=*, mark size=0.6pt] coordinates {(0,42.9) (25,62.9) (50,62.1) (75,62.1) (100,63.6) (125,59.3) (150,66.4)};
\addplot[vanRed, densely dotted, line width=0.5pt] coordinates {(0,43.3) (25,70.1) (50,63.4) (75,64.9) (100,62.7) (125,63.4) (150,67.2)};
\addplot[r2sGreen, line width=0.7pt, mark=*, mark size=0.6pt] coordinates {(0,42.9) (25,75.0) (50,82.1) (75,80.7) (100,83.6) (125,80.7) (150,78.6)};
\addplot[r2sGreen, densely dotted, line width=0.5pt] coordinates {(0,43.3) (25,76.9) (50,83.6) (75,85.1) (100,88.1) (125,82.1) (150,79.9)};
\draw[dashed, black!80, line width=0.6pt] (rel axis cs:0,0.6693) -- (rel axis cs:1,0.6693);
\end{axis}
\end{tikzpicture}
\hfill
\begin{tikzpicture}
\begin{axis}[
  width=1.9cm, height=2.1cm, scale only axis,
  xmin=0, xmax=150, ymin=0, ymax=90, xtick={0,50,100,150},  xlabel={Step},
  axis lines*=left, axis line style={black!45}, tick style={draw=none}, ytick pos=left,
  ymajorgrids, grid style={densely dotted, black!25}, every tick label/.append style={font=\tiny},
  ylabel style={font=\tiny, yshift=-3pt}, xlabel style={font=\tiny, yshift=4pt}, title style={font=\scriptsize, yshift=-3pt},
  title={30B-A3B$\to$4B},
]
\addplot[vanRed, line width=0.7pt, mark=*, mark size=0.6pt] coordinates {(0,29.3) (25,34.3) (50,37.9) (75,33.6) (100,39.3) (125,42.1) (150,41.4)};
\addplot[vanRed, densely dotted, line width=0.5pt] coordinates {(0,35.1) (25,32.1) (50,35.8) (75,41.0) (100,40.3) (125,29.9) (150,37.3)};
\addplot[r2sGreen, line width=0.7pt, mark=*, mark size=0.6pt] coordinates {(0,29.3) (25,34.3) (50,36.4) (75,42.9) (100,43.6) (125,48.6) (150,55.7)};
\addplot[r2sGreen, densely dotted, line width=0.5pt] coordinates {(0,35.1) (25,27.6) (50,32.8) (75,43.3) (100,45.5) (125,48.5) (150,54.5)};
\draw[dashed, black!80, line width=0.6pt] (rel axis cs:0,0.4788) -- (rel axis cs:1,0.4788);
\end{axis}
\end{tikzpicture}
\hfill
\begin{tikzpicture}
\begin{axis}[
  width=1.9cm, height=2.1cm, scale only axis,
  xmin=0, xmax=150, ymin=0, ymax=90, xtick={0,50,100,150},  xlabel={Step},
  axis lines*=left, axis line style={black!45}, tick style={draw=none}, ytick pos=left,
  ymajorgrids, grid style={densely dotted, black!25}, every tick label/.append style={font=\tiny},
  ylabel style={font=\tiny, yshift=-3pt}, xlabel style={font=\tiny, yshift=4pt}, title style={font=\scriptsize, yshift=-3pt},
  title={30B-A3B$\to$8B},
]
\addplot[vanRed, line width=0.7pt, mark=*, mark size=0.6pt] coordinates {(0,42.9) (25,44.3) (50,41.4) (75,46.4) (100,47.9) (125,48.6) (150,45.7)};
\addplot[vanRed, densely dotted, line width=0.5pt] coordinates {(0,50.7) (25,43.3) (50,44.8) (75,50.0) (100,52.2) (125,37.3) (150,39.6)};
\addplot[r2sGreen, line width=0.7pt, mark=*, mark size=0.6pt] coordinates {(0,42.9) (25,40.7) (50,48.6) (75,53.6) (100,52.9) (125,54.3) (150,57.1)};
\addplot[r2sGreen, densely dotted, line width=0.5pt] coordinates {(0,50.7) (25,50.0) (50,51.5) (75,53.7) (100,53.7) (125,51.5) (150,54.5)};
\draw[dashed, black!80, line width=0.6pt] (rel axis cs:0,0.4788) -- (rel axis cs:1,0.4788);
\end{axis}
\end{tikzpicture}
\hfill
\\[3pt]
\makebox[\linewidth][c]{\textbf{WebShop}}\\[1pt]
\begin{tikzpicture}
\begin{axis}[
  width=1.9cm, height=2.1cm, scale only axis,
  xmin=0, xmax=150, ymin=0, ymax=65, xtick={0,50,100,150}, ylabel={Success}, xlabel={Step},
  axis lines*=left, axis line style={black!45}, tick style={draw=none}, ytick pos=left,
  ymajorgrids, grid style={densely dotted, black!25}, every tick label/.append style={font=\tiny},
  ylabel style={font=\tiny, yshift=-3pt}, xlabel style={font=\tiny, yshift=4pt}, title style={font=\scriptsize, yshift=-3pt},
  title={9B$\to$2B},
]
\addplot[vanRed, line width=0.7pt, mark=*, mark size=0.6pt] coordinates {(0,18.0) (25,29.0) (50,26.4) (75,35.2) (100,35.0) (125,34.2) (150,37.0)};
\addplot[vanRed, densely dotted, line width=0.5pt] coordinates {(0,13.7) (25,30.1) (50,25.4) (75,34.8) (100,30.9) (125,33.2) (150,33.2)};
\addplot[r2sGreen, line width=0.7pt, mark=*, mark size=0.6pt] coordinates {(0,18.0) (25,44.0) (50,57.4) (75,55.4) (100,54.4) (125,56.4) (150,56.2)};
\addplot[r2sGreen, densely dotted, line width=0.5pt] coordinates {(0,13.7) (25,40.6) (50,51.2) (75,53.9) (100,51.2) (125,54.7) (150,52.3)};
\draw[dashed, black!80, line width=0.6pt] (rel axis cs:0,0.5159) -- (rel axis cs:1,0.5159);
\end{axis}
\end{tikzpicture}
\hfill
\begin{tikzpicture}
\begin{axis}[
  width=1.9cm, height=2.1cm, scale only axis,
  xmin=0, xmax=150, ymin=0, ymax=65, xtick={0,50,100,150},  xlabel={Step},
  axis lines*=left, axis line style={black!45}, tick style={draw=none}, ytick pos=left,
  ymajorgrids, grid style={densely dotted, black!25}, every tick label/.append style={font=\tiny},
  ylabel style={font=\tiny, yshift=-3pt}, xlabel style={font=\tiny, yshift=4pt}, title style={font=\scriptsize, yshift=-3pt},
  title={9B$\to$4B},
]
\addplot[vanRed, line width=0.7pt, mark=*, mark size=0.6pt] coordinates {(0,33.2) (25,37.2) (50,36.4) (75,36.2) (100,38.2) (125,35.6) (150,37.8)};
\addplot[vanRed, densely dotted, line width=0.5pt] coordinates {(0,30.5) (25,34.4) (50,32.0) (75,32.8) (100,30.9) (125,27.0) (150,32.4)};
\addplot[r2sGreen, line width=0.7pt, mark=*, mark size=0.6pt] coordinates {(0,33.2) (25,55.8) (50,54.8) (75,56.4) (100,56.8) (125,57.8) (150,54.2)};
\addplot[r2sGreen, densely dotted, line width=0.5pt] coordinates {(0,30.5) (25,52.7) (50,53.5) (75,52.3) (100,51.6) (125,53.9) (150,55.5)};
\draw[dashed, black!80, line width=0.6pt] (rel axis cs:0,0.5159) -- (rel axis cs:1,0.5159);
\end{axis}
\end{tikzpicture}
\hfill
\begin{tikzpicture}
\begin{axis}[
  width=1.9cm, height=2.1cm, scale only axis,
  xmin=0, xmax=150, ymin=0, ymax=65, xtick={0,50,100,150},  xlabel={Step},
  axis lines*=left, axis line style={black!45}, tick style={draw=none}, ytick pos=left,
  ymajorgrids, grid style={densely dotted, black!25}, every tick label/.append style={font=\tiny},
  ylabel style={font=\tiny, yshift=-3pt}, xlabel style={font=\tiny, yshift=4pt}, title style={font=\scriptsize, yshift=-3pt},
  title={30B-A3B$\to$4B},
]
\addplot[vanRed, line width=0.7pt, mark=*, mark size=0.6pt] coordinates {(0,15.2) (25,25.4) (50,30.8) (75,21.2) (100,25.6) (125,21.8) (150,23.6)};
\addplot[vanRed, densely dotted, line width=0.5pt] coordinates {(0,13.7) (25,23.0) (50,27.3) (75,17.6) (100,20.7) (125,20.3) (150,19.9)};
\addplot[r2sGreen, line width=0.7pt, mark=*, mark size=0.6pt] coordinates {(0,15.2) (25,28.2) (50,36.2) (75,25.8) (100,22.0) (125,28.0) (150,28.0)};
\addplot[r2sGreen, densely dotted, line width=0.5pt] coordinates {(0,13.7) (25,24.2) (50,29.7) (75,28.1) (100,22.7) (125,27.3) (150,27.3)};
\draw[dashed, black!80, line width=0.6pt] (rel axis cs:0,0.0441) -- (rel axis cs:1,0.0441);
\end{axis}
\end{tikzpicture}
\hfill
\begin{tikzpicture}
\begin{axis}[
  width=1.9cm, height=2.1cm, scale only axis,
  xmin=0, xmax=150, ymin=0, ymax=65, xtick={0,50,100,150},  xlabel={Step},
  axis lines*=left, axis line style={black!45}, tick style={draw=none}, ytick pos=left,
  ymajorgrids, grid style={densely dotted, black!25}, every tick label/.append style={font=\tiny},
  ylabel style={font=\tiny, yshift=-3pt}, xlabel style={font=\tiny, yshift=4pt}, title style={font=\scriptsize, yshift=-3pt},
  title={30B-A3B$\to$8B},
]
\addplot[vanRed, line width=0.7pt, mark=*, mark size=0.6pt] coordinates {(0,22.2) (25,18.6) (50,24.6) (75,22.4) (100,24.4) (125,22.4) (150,20.4)};
\addplot[vanRed, densely dotted, line width=0.5pt] coordinates {(0,20.3) (25,13.3) (50,19.9) (75,19.5) (100,22.7) (125,19.1) (150,19.9)};
\addplot[r2sGreen, line width=0.7pt, mark=*, mark size=0.6pt] coordinates {(0,22.2) (25,18.2) (50,32.8) (75,30.8) (100,34.4) (125,26.8) (150,31.4)};
\addplot[r2sGreen, densely dotted, line width=0.5pt] coordinates {(0,20.3) (25,13.3) (50,25.8) (75,23.8) (100,32.4) (125,26.6) (150,30.1)};
\draw[dashed, black!80, line width=0.6pt] (rel axis cs:0,0.0441) -- (rel axis cs:1,0.0441);
\end{axis}
\end{tikzpicture}
\hfill
\\[3pt]
\makebox[\linewidth][c]{\textbf{ScienceWorld}}\\[1pt]
\begin{tikzpicture}
\begin{axis}[
  width=1.9cm, height=2.1cm, scale only axis,
  xmin=0, xmax=150, ymin=0, ymax=60, xtick={0,50,100,150}, ylabel={Success}, xlabel={Step},
  axis lines*=left, axis line style={black!45}, tick style={draw=none}, ytick pos=left,
  ymajorgrids, grid style={densely dotted, black!25}, every tick label/.append style={font=\tiny},
  ylabel style={font=\tiny, yshift=-3pt}, xlabel style={font=\tiny, yshift=4pt}, title style={font=\scriptsize, yshift=-3pt},
  title={9B$\to$2B},
]
\addplot[vanRed, line width=0.7pt, mark=*, mark size=0.6pt] coordinates {(0,2.9) (25,14.3) (50,16.4) (75,20.0) (100,25.0) (125,31.4) (150,24.3)};
\addplot[vanRed, densely dotted, line width=0.5pt] coordinates {(0,4.5) (25,16.4) (50,20.9) (75,25.4) (100,20.9) (125,23.9) (150,24.6)};
\addplot[r2sGreen, line width=0.7pt, mark=*, mark size=0.6pt] coordinates {(0,2.9) (25,22.9) (50,29.3) (75,35.0) (100,36.4) (125,44.3) (150,40.0)};
\addplot[r2sGreen, densely dotted, line width=0.5pt] coordinates {(0,4.5) (25,23.9) (50,28.4) (75,25.4) (100,29.9) (125,30.6) (150,34.3)};
\draw[dashed, black!80, line width=0.6pt] (rel axis cs:0,0.5595) -- (rel axis cs:1,0.5595);
\end{axis}
\end{tikzpicture}
\hfill
\begin{tikzpicture}
\begin{axis}[
  width=1.9cm, height=2.1cm, scale only axis,
  xmin=0, xmax=150, ymin=0, ymax=60, xtick={0,50,100,150},  xlabel={Step},
  axis lines*=left, axis line style={black!45}, tick style={draw=none}, ytick pos=left,
  ymajorgrids, grid style={densely dotted, black!25}, every tick label/.append style={font=\tiny},
  ylabel style={font=\tiny, yshift=-3pt}, xlabel style={font=\tiny, yshift=4pt}, title style={font=\scriptsize, yshift=-3pt},
  title={9B$\to$4B},
]
\addplot[vanRed, line width=0.7pt, mark=*, mark size=0.6pt] coordinates {(0,32.1) (25,39.3) (50,34.3) (75,20.7) (100,24.3) (125,30.7) (150,40.7)};
\addplot[vanRed, densely dotted, line width=0.5pt] coordinates {(0,31.3) (25,30.6) (50,32.1) (75,24.6) (100,29.1) (125,32.1) (150,35.8)};
\addplot[r2sGreen, line width=0.7pt, mark=*, mark size=0.6pt] coordinates {(0,32.1) (25,42.9) (50,47.9) (75,47.9) (100,50.0) (125,53.6) (150,54.3)};
\addplot[r2sGreen, densely dotted, line width=0.5pt] coordinates {(0,31.3) (25,29.1) (50,43.3) (75,41.8) (100,38.8) (125,41.8) (150,46.3)};
\draw[dashed, black!80, line width=0.6pt] (rel axis cs:0,0.5595) -- (rel axis cs:1,0.5595);
\end{axis}
\end{tikzpicture}
\hfill
\begin{tikzpicture}
\begin{axis}[
  width=1.9cm, height=2.1cm, scale only axis,
  xmin=0, xmax=150, ymin=0, ymax=60, xtick={0,50,100,150},  xlabel={Step},
  axis lines*=left, axis line style={black!45}, tick style={draw=none}, ytick pos=left,
  ymajorgrids, grid style={densely dotted, black!25}, every tick label/.append style={font=\tiny},
  ylabel style={font=\tiny, yshift=-3pt}, xlabel style={font=\tiny, yshift=4pt}, title style={font=\scriptsize, yshift=-3pt},
  title={30B-A3B$\to$4B},
]
\addplot[vanRed, line width=0.7pt, mark=*, mark size=0.6pt] coordinates {(0,8.6) (25,20.7) (50,32.9) (75,26.4) (100,36.4) (125,35.7) (150,39.3)};
\addplot[vanRed, densely dotted, line width=0.5pt] coordinates {(0,9.7) (25,29.1) (50,35.8) (75,38.8) (100,42.5) (125,36.6) (150,38.1)};
\addplot[r2sGreen, line width=0.7pt, mark=*, mark size=0.6pt] coordinates {(0,8.6) (25,25.0) (50,43.6) (75,44.3) (100,55.0) (125,55.0) (150,48.6)};
\addplot[r2sGreen, densely dotted, line width=0.5pt] coordinates {(0,9.7) (25,36.6) (50,39.6) (75,44.0) (100,46.3) (125,47.8) (150,41.8)};
\draw[dashed, black!80, line width=0.6pt] (rel axis cs:0,0.4921) -- (rel axis cs:1,0.4921);
\end{axis}
\end{tikzpicture}
\hfill
\begin{tikzpicture}
\begin{axis}[
  width=1.9cm, height=2.1cm, scale only axis,
  xmin=0, xmax=150, ymin=0, ymax=60, xtick={0,50,100,150},  xlabel={Step},
  axis lines*=left, axis line style={black!45}, tick style={draw=none}, ytick pos=left,
  ymajorgrids, grid style={densely dotted, black!25}, every tick label/.append style={font=\tiny},
  ylabel style={font=\tiny, yshift=-3pt}, xlabel style={font=\tiny, yshift=4pt}, title style={font=\scriptsize, yshift=-3pt},
  title={30B-A3B$\to$8B},
]
\addplot[vanRed, line width=0.7pt, mark=*, mark size=0.6pt] coordinates {(0,20.7) (25,33.6) (50,39.3) (75,42.1) (100,39.3) (125,42.1) (150,40.0)};
\addplot[vanRed, densely dotted, line width=0.5pt] coordinates {(0,23.1) (25,33.6) (50,41.0) (75,38.1) (100,33.6) (125,41.0) (150,34.3)};
\addplot[r2sGreen, line width=0.7pt, mark=*, mark size=0.6pt] coordinates {(0,20.7) (25,36.4) (50,46.4) (75,56.4) (100,55.7) (125,57.9) (150,56.4)};
\addplot[r2sGreen, densely dotted, line width=0.5pt] coordinates {(0,23.1) (25,29.9) (50,41.0) (75,41.0) (100,45.5) (125,42.5) (150,43.3)};
\draw[dashed, black!80, line width=0.6pt] (rel axis cs:0,0.4921) -- (rel axis cs:1,0.4921);
\end{axis}
\end{tikzpicture}
\hfill
\\[3pt]
{\tiny \tikz[baseline=-0.4ex]{\draw[vanRed, line width=0.7pt] (0,0.065) -- (0.4,0.065);}\,Vanilla OPD\quad \tikz[baseline=-0.4ex]{\draw[r2sGreen, line width=0.7pt] (0,0.065) -- (0.4,0.065);}\,\method{}\quad Solid: seen / canonical\quad Dotted: unseen / pool\quad \tikz[baseline=-0.4ex]{\draw[dashed, black!80, line width=0.6pt] (0,0.065) -- (0.4,0.065);}\,Teacher Alone}
\caption{\textbf{Success over training in every cell of \Cref{tab:main}.} Vanilla and \method{} students, one decode every 25 updates (update 0: untrained). Solid: seen split or WebShop's canonical goals; dotted: unseen split or its 256-goal pool; dashed: the teacher acting alone on the seen split.}
\label{fig:curves-all}
\end{figure}

\textbf{Training curves complement the final-checkpoint comparison.} \Cref{fig:curves-all} plots the in-training evaluation of the vanilla and \method{} students of every cell of \Cref{tab:main}: one decode every 25 updates under the training seed, so the final point of each curve is the first of the three decodes behind the table. The \method{} student ends above vanilla OPD in all twelve cells, on both splits, and above the teacher acting alone on the seen split in all twelve, with no checkpoint selection.

\textbf{Three training runs per method.} \Cref{fig:teacher-bars} follows vanilla OPD, TCOD-F2B, Guided-OPD, FutureBridge and \method{} from the Qwen3.5-9B teacher to the Qwen3.5-4B student under the protocol of \Cref{app:protocol}. Each method has three training runs per environment, started from the same untrained student and differing only in the training seed; the three \method{} runs share the one repaired teacher of \Cref{tab:main}, so its band reflects student training alone. Each run is evaluated every 25 updates under its own seed, update 0 being the untrained student common to all five curves. Lines are means over the three runs and bands one sample standard deviation, without interpolation or checkpoint selection; the dashed line is the teacher acting alone.

\end{document}